%% file: example_paper.tex
\documentclass[nohyperref]{article}

\usepackage{microtype}
\usepackage{graphicx}
\usepackage{subcaption}
\usepackage{caption}
\usepackage{booktabs} 
\usepackage{algorithm,algpseudocode}

\usepackage{hyperref}

\usepackage{bbm}

\usepackage{xcolor}

\usepackage[accepted]{icml2022}

\usepackage{amsmath}
\usepackage{amssymb}
\usepackage{mathtools}
\usepackage{amsthm}

\usepackage[capitalize,noabbrev]{cleveref}

\theoremstyle{plain}

\theoremstyle{definition}

\theoremstyle{remark}

\usepackage[textsize=tiny]{todonotes}

\icmltitlerunning{Improving Policy Optimization with Generalist-Specialist Learning}

\begin{document}

\twocolumn[
\icmltitle{Improving Policy Optimization with Generalist-Specialist Learning}




\icmlsetsymbol{equal}{*}

\begin{icmlauthorlist}
\icmlauthor{Zhiwei Jia}{ucsd}
\icmlauthor{Xuanlin Li}{ucsd}
\icmlauthor{Zhan Ling}{ucsd}
\icmlauthor{Shuang Liu}{ucsd}
\icmlauthor{Yiran Wu}{ucsd}
\icmlauthor{Hao Su}{ucsd}
\end{icmlauthorlist}

\icmlaffiliation{ucsd}{University of California, San Diego}

\icmlcorrespondingauthor{Zhiwei Jia}{zjia@eng.ucsd.edu}
\icmlcorrespondingauthor{Hao Su}{haosu@eng.ucsd.edu}

\icmlkeywords{Policy Optimization, Reinforcement Learning, Machine Learning, ICML}

\vskip 0.3in
]

\newcommand{\hao}[1]{\textcolor{red}{[hao: #1]}}



\printAffiliationsAndNotice{} 

\begin{abstract}
Generalization in deep reinforcement learning over unseen environment variations usually requires policy learning over a large set of diverse training variations. 
We empirically observe that an agent trained on many variations (a \emph{generalist}) tends to learn faster at the beginning, yet its performance plateaus at a less optimal level for a long time. In contrast, an agent trained only on a few variations (a \emph{specialist}) can often achieve high returns under a limited computational budget. To have the best of both worlds, we propose a novel generalist-specialist training framework. Specifically, we first train a generalist on all environment variations; when it fails to improve, we launch a large population of specialists with weights cloned from the generalist, each trained to master a selected small subset of variations. We finally resume the training of the generalist with auxiliary rewards induced by demonstrations of all specialists. In particular, we investigate the timing to start specialist training and compare strategies to learn generalists with assistance from specialists.
We show that this framework pushes the envelope of policy learning on several challenging and popular benchmarks including Procgen, Meta-World and ManiSkill.
\end{abstract}


\input{sections/introduction}

\input{sections/related_work}
\input{sections/background}

\input{sections/illustrative_exp}
\input{sections/method}

\input{sections/experiments}
\input{sections/ablations}

\input{sections/conclusion}

\bibliography{example_paper}
\bibliographystyle{icml2022}

\newpage
\appendix
\onecolumn

\input{sections/appendix}

\end{document}

%% file: sections/introduction.tex
\section{Introduction}

Deep Reinforcement Learning (DRL) holds the promise in a wide range of applications such as autonomous vehicles \cite{filos2020can}, gaming \cite{silver2017mastering}, robotics \cite{kalashnikov2018qt} and healthcare \cite{yu2021reinforcement}. 
To fulfill the potential of RL, we need algorithms and models capable of adapting and generalizing to unseen (but similar) environment variations during their deployment.
Recently, several benchmarks \cite{gupta2018robot,cobbe2020leveraging,mu2021maniskill,szot2021habitat,gan2020threedworld,zhao2021luminous} were proposed to this end, featuring a very high diversity of variations in training environments, accomplished through procedural generation and layout randomization, to encourage policy generalization.

However, due to the sheer number of variations, many existing DRL algorithms struggle to efficiently achieve high performance during training, let alone generalization.
For instance, in Procgen Benchmark \cite{cobbe2020leveraging}, a PPO \cite{schulman2017proximal} agent trained on a thousand levels can have poor performance even with hundreds of millions of samples.
Training PPO on visual navigation tasks involving $\sim$100 scenes might require billions of samples to achieve good performance \cite{wijmans2019dd}.
Several lines of work have been proposed to alleviate this issue, by accelerating training with automatic curriculum \cite{jiang2021prioritized}, improving learned representations with the help of extra constraints \cite{igl2019generalization,raileanu2020automatic}, or decoupling the learning of the policy and value networks \cite{cobbe2021phasic,raileanu2021decoupling}.



Orthogonal to these approaches, we tackle the challenge from a perspective inspired by how human organizations solve difficult problems. We first define a \emph{generalist} agent to be a single policy that can solve all environment variations. We also define a \emph{specialist} agent to be a policy that masters a subset, but not all, of environment variations.
Our goal is to utilize experiences from the \emph{specialists} to aid the policy optimization of the \emph{generalist}. 

We observe that trajectories belonging to different environment variations often consist of shared early stages and context-specific later stages. It is often more efficient for a single generalist to learn the \emph{shared} early stages for all contexts than first training different specialists and then merging them to a generalist. For instance, learning to push a chair towards a goal, regardless of variations in chair geometry, topology, or dynamics, requires an agent to first recognize the chair and approach it. During these early stages, jointly training an agent on all chairs, starting and goal states results in faster learning. However, as the visited states get more and more diverse, it becomes increasingly hard for the policy and value network to maintain the predictive power without forgetting (i.e., ``catastrophic forgetting''), or be vigilant to input dimensions that were not useful at the beginning but crucial for later stages (i.e., ``catastrophic ignorance'' in Sec.~\ref{sec:illustrative_example}). 
This poses a significant challenge and results in performance plateaus. 
Meanwhile, if we only consider a small subset of environment variations and train a specialist on it, then due to the low state variance, the specialist can often master these variations, achieving a higher return on them than the generalist.

\begin{figure}[t]
\centering
\includegraphics[width=0.95\linewidth]{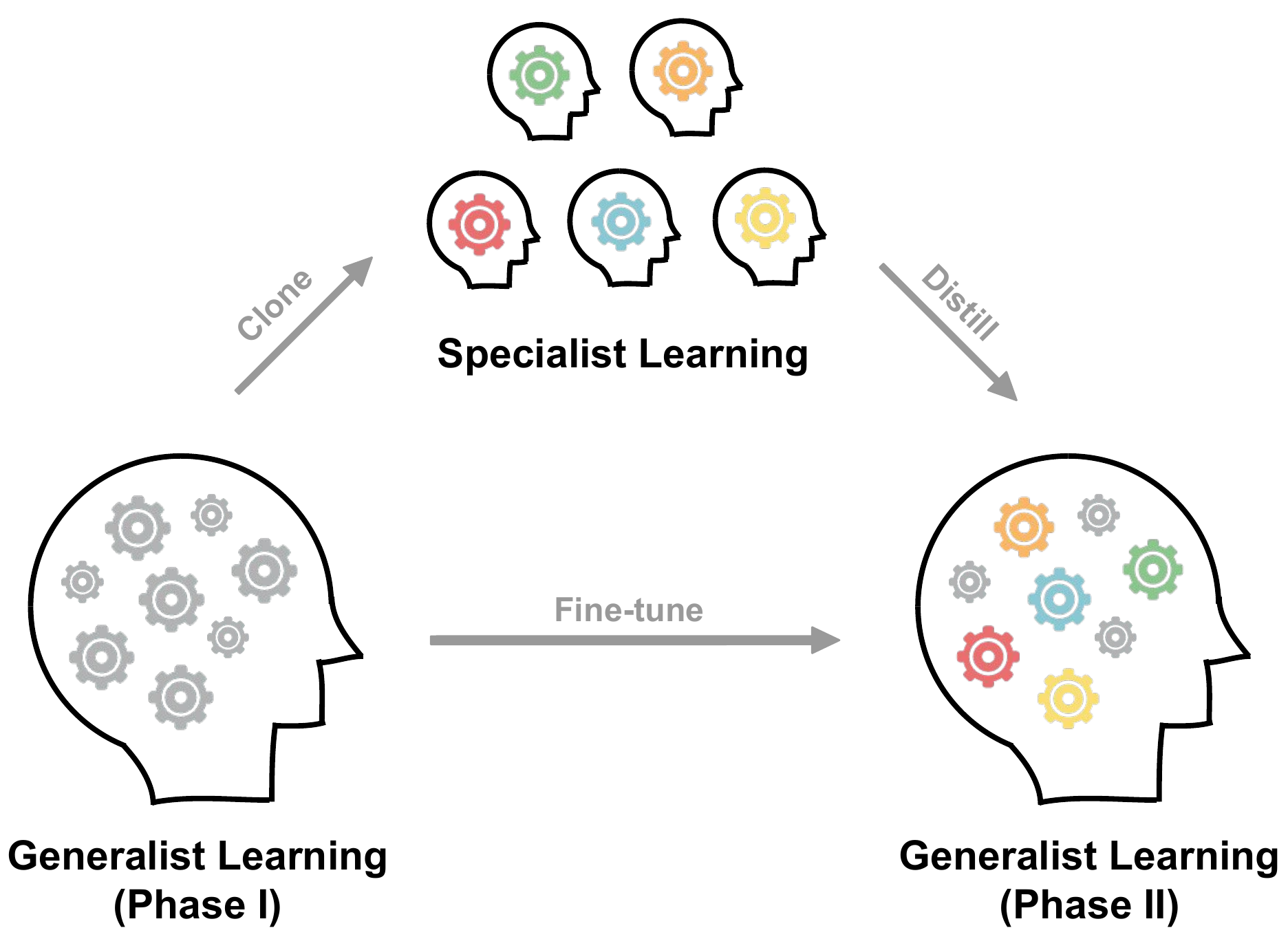}
\vspace{-1mm}
\caption{Illustration of Generalist-Specialist Learning (GSL) framework. During initial generalist learning (phase I), a generalist agent learns to master all environment variations at once. Next, each specialist agent works on a subset of environment variations. Finally, the generalist is fine-tuned (phase II) with guidance provided by the specialists (e.g., via demonstrations).}
\label{fig:teaser}
\end{figure}

Inspired by these observations, we propose a novel framework\footnote{As a meta-algorithm, the (pseudo)code is available \href{https://github.com/SeanJia/GSL}{here}.}, \textbf{Generalist-Specialist Learning (GSL)}, to take advantage of both \emph{specialist}'s high return and \emph{generalist}'s faster early training. As illustrated in Fig. \ref{fig:teaser}, we first optimize the generalist on all environment variations for fast initial policy learning. When it fails to improve (by a simple criteria), we launch a large population of specialist agents, each loaded from the generalist checkpoint, and optimize on a small subset of environment variations. After specialists' performance quickly surpass the generalist's, we use specialists' demonstrations as auxiliary rewards for generalist training, advancing its own performance. While some previous approaches also involve specialist training \cite{mu2021maniskill,teh2017distral,ghosh2017divide,mu2020refactoring,chen2022system}, they either do not realize or diagnose the gains and losses for generalist vs. specialist, or focus on a different setup with their essential idea orthogonal to our proposal. We demonstrate the effectiveness of our framework on subsets of two very challenging benchmarks: Procgen \cite{procgen} that consists of procedurally generated 2D games (with 1024 training levels, more than the official 200 levels setup), and ManiSkill \cite{mu2021maniskill} that evaluates physical manipulation skills over diverse 3D objects with high geometry and topology variations.

%% file: sections/related_work.tex
\section{Related Work}

\paragraph{Divide-and-Conquer in RL}
Our work is most closely related to works along this line. Previous works like \citep{teh2017distral, ghosh2017divide} have adopted divide-and-conquer for training an RL agent. They split the state space into subsets, and alternate between training each local policy on each subset and merging the local policies into a single center policy using imitation learning. Although these approaches also adopt a perspective of generalists and specialists, they did not study the timing for starting specialist training, which is a key contribution of our work. In addition, when distilling experiences from specialists into a generalist, they usually use behavior cloning and address the demonstration inconsistency issue by KL divergence regularization between local policies \cite{ghosh2017divide}, which requires a synchronized training strategy, making it intractable when scaling to large numbers of specialists as in our experiments. 
We show that our approach can still work well with large number of specialists.

\paragraph{Policy Distillation and Imitation Learning}
Distilling a single generalist (student) from a group of specialists (teachers) is a promising way to achieve good performance on challenging tasks ~\citep{rusu2015policy, ross2011reduction, mu2020refactoring,czarnecki2018mix,team2021open}. Population-based agent training (PBT) was utilized in \cite{czarnecki2018mix,team2021open}. 
Previous works like \cite{rusu2015policy} have adopted supervised learning to training a generalist over specialists' demonstrations. Other works convert demonstrations into rewards for online learning \cite{finn2016guided, gail,irl,dapg,shen2022learning}. 
Our framework makes good use of the policy distillation as its sub-module.

\paragraph{Large-Scale RL}
Training an RL agent over a large number of environment variations is a promising approach to obtaining a generalizable policy ~\citep{procgen,mu2021maniskill}. One series of benchmarks for this objective involve variations on object geometry and topology ~\citep{mu2021maniskill} and task semantics ~\citep{yu2020meta, james2020rlbench}, which are usually mentioned as multi-task RL benchmarks. Another series of benchmarks procedurally generate diverse levels and layouts for an environment ~\citep{urakami2019doorgym, procgen}. For both series of benchmarks, training a single agent over multiple variations is known to be challenging, which warrants more exploration into this field. 






\paragraph{Multi-task RL}
In multi-task RL, an agent is trained on multiple tasks given a task-specific encoding. Recent works have made significant progress accelerating policy optimization across multiple tasks. One stream of work focuses on improving task (context) encoding representations from environment dynamics or reward signal \cite{bram2019attentive,sodhani2021multi}.
Another stream focuses on studying and alleviating negative gradient interference from different tasks \cite{schaul2019ray,yu2020gradient,kurin2022defense}.
Different from these approaches, our framework is designed for general RL tasks, which not only encompasses multi-task RL environments, but also general RL environments such as Procgen, whose environment variations do not have semantic encoding (i.e, each variation is represented by a random seed). 








%% file: sections/background.tex
\section{Background}

A general \textbf{Markov decision Process} (MDP) is defined as a tuple $M=(S, A, T, R, \gamma)$, where $S, A$ are state space and action space, $T(s'|s, a)$ is the state transition probability, $R(s, a)$ is the reward function, and $\gamma \in [0, 1)$ is the discount factor. In reinforcement learning, we aim to train a policy $\pi(a|s)$ that maximizes the expected accumulated return given by $J(\pi)=E_{(s_t, a_t) \sim \rho_{\pi}}[\sum_{t=0} \gamma^t r(s_t, a_t)]$.

A \textbf{Block Contextual Markov Decision Process} (BC-MDP) ~\citep{zhang2020invariant, du2019provably, pmlr-v139-sodhani21a} augments an MDP with context information, which can be defined as $(C, A, M(c), \gamma)$, where $C$ is the context space, $M$ is a function which maps any context $c \in C$ to an MDP $M(c) = \{S^c, T^c, R^c\}$. BC-MDP can be adapted to the multi-task setting, where contexts control objects used in the environments (e.g. object variations in ManiSkill~\citep{mu2021maniskill}) and task semantics (e.g. tasks in MetaWorld~\citep{yu2020meta}). BC-MDP can also be adapted to a general MDP to control the random seeds (e.g. seeds for procedural generation in Procgen~\citep{procgen}). In this paper, we consider both settings. Unlike previous works~\citep{pmlr-v139-sodhani21a}, our framework does not require the context to contain any semantic meaning, which is typical in the multi-task RL setting. The context can be anything that splits an MDP into several sub-MDPs.

In the following sections, we first motivate our approach by analyzing a simple example in Section \ref{sec:illustrative_example}. We then introduce several important ingredients in our Generalist-Specialist Learning (GSL) framework.


%% file: sections/illustrative_exp.tex
\section{An Illustrative Example}
\label{sec:illustrative_example}

\begin{figure}[h]
\centering
\includegraphics[width=0.68\linewidth]{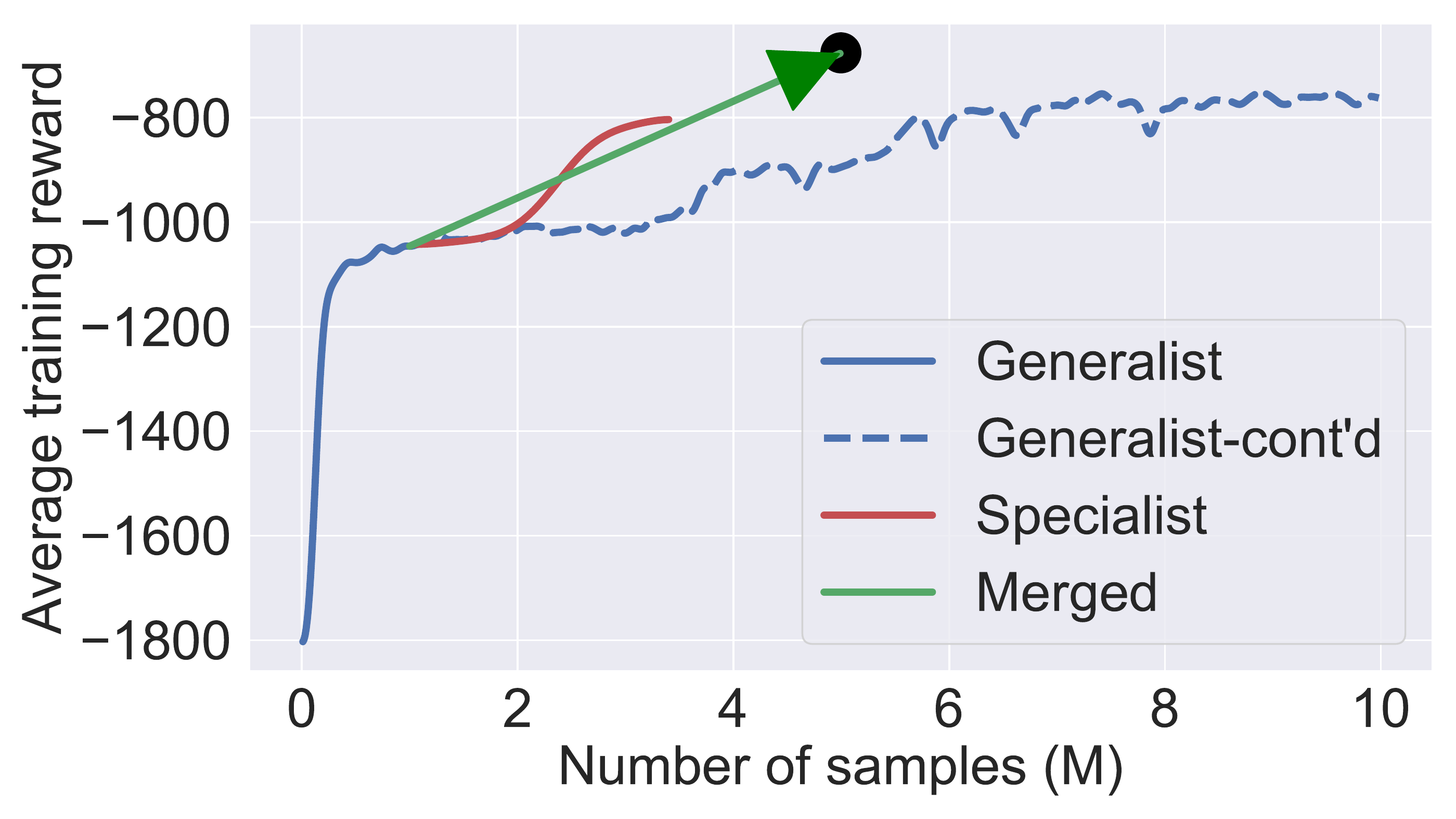}
\vspace{-1mm}
\caption{Training curve of PPO on the brush-like maze. The generalist learns fast but plateaus quickly (dashed blue line). The specialists (cloned from the generalist) learn to solve individual goals better (solid red line). The fine-tuned generalist (trained by DAPG) with the specialists' demos achieves the best results efficiently (green arrow). This improvement is consistent among 5 runs. See details in Appendix.}
\label{fig:ill-res}
\end{figure}

\begin{figure*}[t]
\centering
\begin{subfigure}[b]{0.245\linewidth}
    \includegraphics[width=\linewidth]{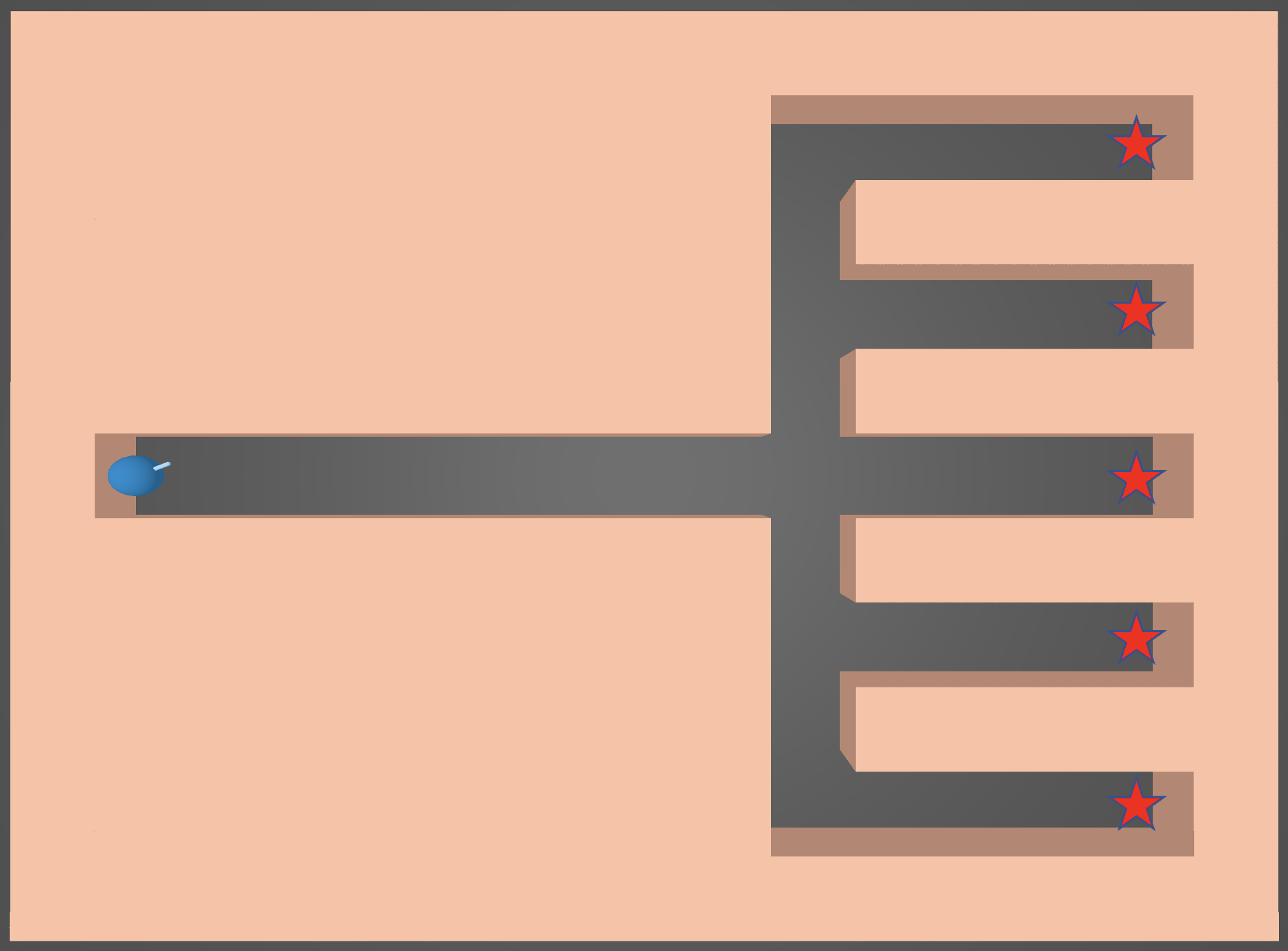}
    \caption{Illustrative environment}
    \label{fig:ill-env}
\end{subfigure}
\begin{subfigure}[b]{0.183\textwidth}
    \begin{subfigure}{\linewidth}
        \centering
        \includegraphics[width=0.48\linewidth]{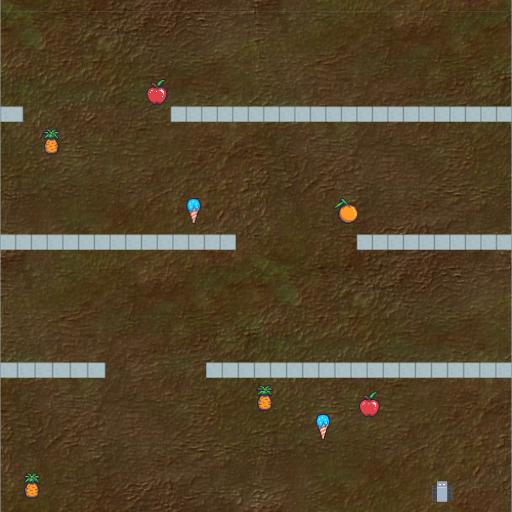}
        \includegraphics[width=0.48\linewidth]{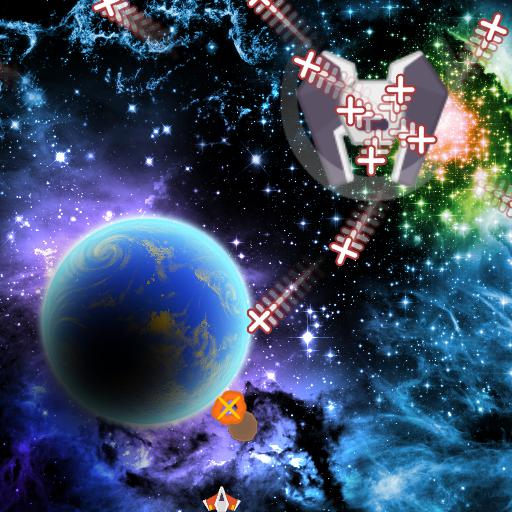}
        \end{subfigure}
    \begin{subfigure}{\linewidth}
        \centering
        \vspace*{.5mm}
        \includegraphics[width=0.48\linewidth]{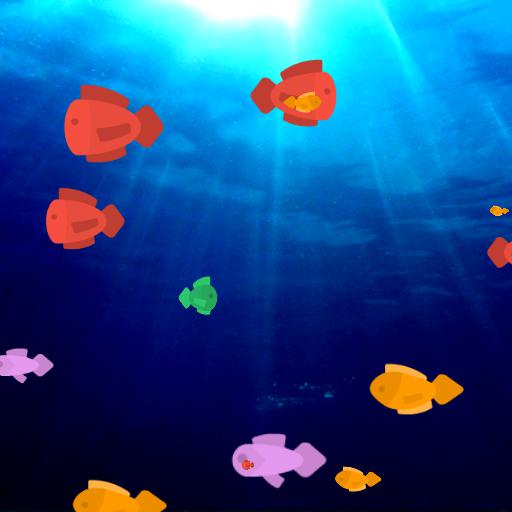}
        \includegraphics[width=0.48\linewidth]{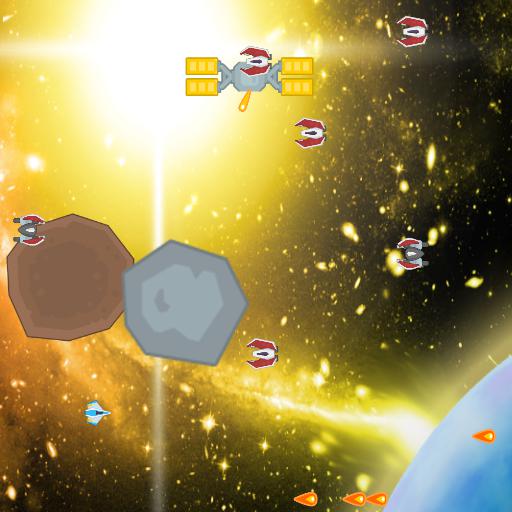}
    \end{subfigure}
    \caption{Procgen}
    \label{fig:prcgen-env}
\end{subfigure}
\begin{subfigure}[b]{0.201\linewidth}
    \includegraphics[width=\linewidth]{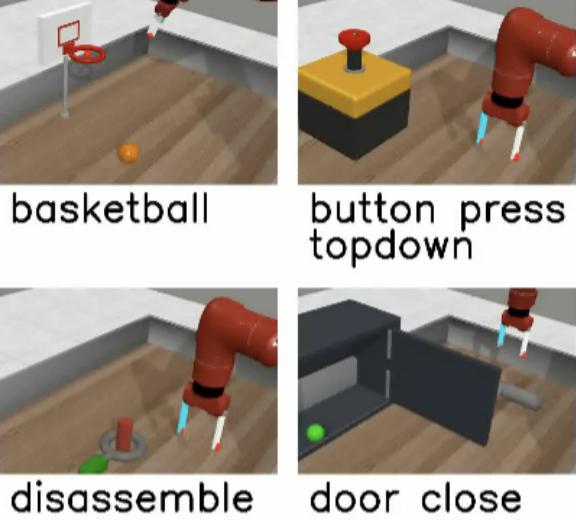}
    \caption{Meta-World}
    \label{fig:metaworld-env}
\end{subfigure}
\begin{subfigure}[b]{0.305\linewidth}
    \includegraphics[width=\linewidth]{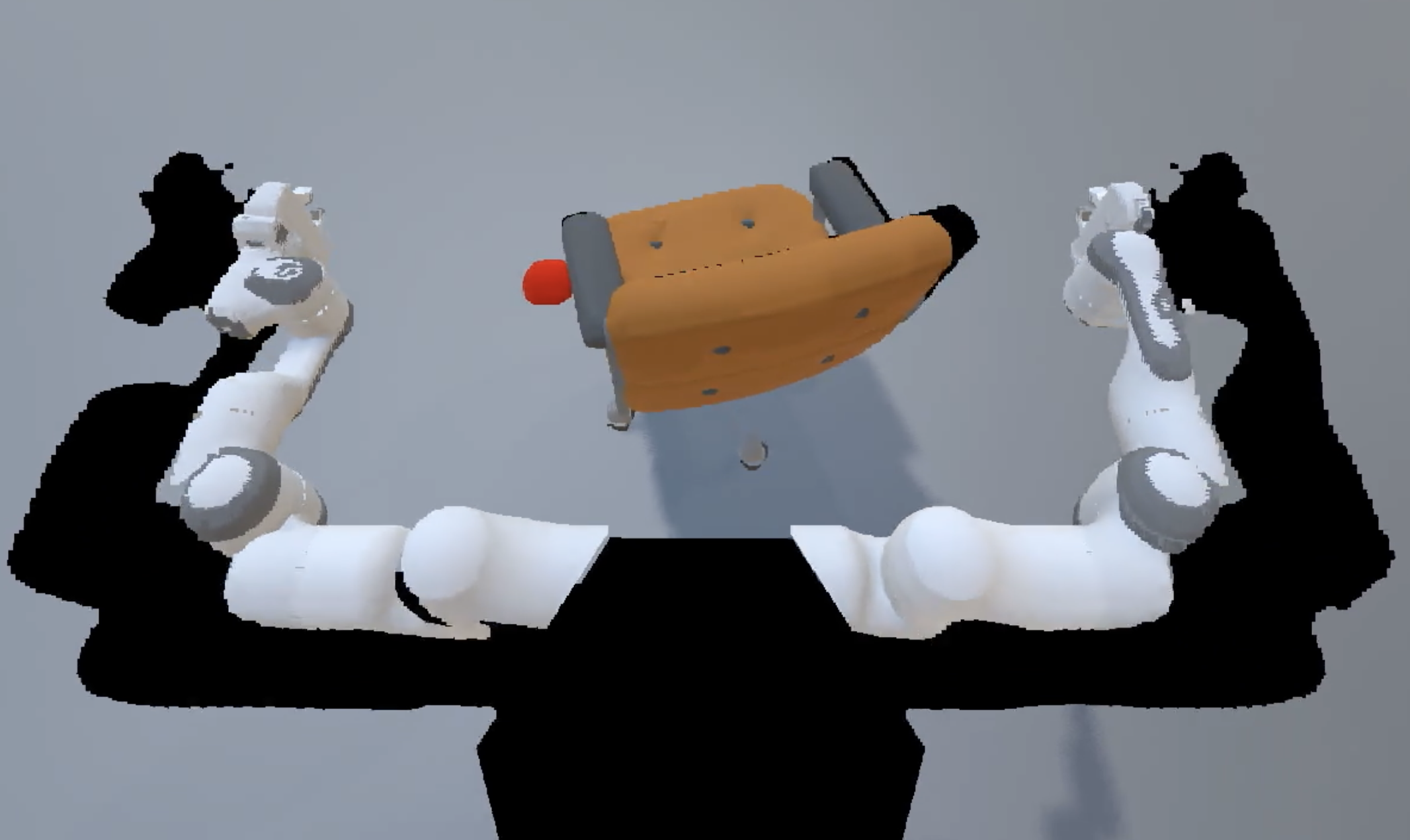}
    \caption{ManiSkill}
    \label{fig:maniskill-env}
\end{subfigure}
\vspace{-1mm}
\caption{(a) An illustrative environment that consists of a long corridor and a ``brush"-like maze, where the agent starts from the left and needs to reach one of the five targets (stars in the figure) specified by a goal context. (b) Sample environments in Procgen, a 2D vision-based game environment where levels are generated procedurally. (c) Sample environments in Meta-World, where the agent is asked to perform diverse manipulation tasks given state-space observations. (d) The PushChair task in ManiSkill, where an agent is required to push a chair towards a goal (red ball) given point cloud observations of diverse chairs. }
\label{fig:environments}
\end{figure*}

Consider a simple ``brush''-like maze illustrated in Fig. \ref{fig:ill-env}. An agent (a \emph{generalist}) starts from the leftmost position in the corridor and needs to reach the goal specified by a real context $c$. For each $i=\{1, 2, \cdots, 5\}$, $C_i=(\frac{i-1}{5}, \frac{i}{5}]$ indicates the context space for the $i$-th goal (marked as red stars). Upon environment resets, we first uniformly sample the goal $i$ and then sample $c$ uniformly from $C_i$. An agent is given the position and velocity (both are real vectors) as an observation and is capable of setting its velocity as an action. The reward is the approximated negative geodesic distance between the current position and the goal.

We train a PPO agent and discover the phenomenon of ``\textit{catastrophic ignorance}'' (we coin this term inspired by ``catastrophic forgetting''). As in Fig.~\ref{fig:ill-res}, in early stages, the agent quickly learns to move rightwards in the maze. However, the agent tends to ignore the goal context $c$ since it plays little role in the early stages. This is revealed by the agent's tendency to always move rightwards after the intersection and arrive at the middle goal regardless of the goal context. It requires a large amount of samples for the agent to learn to extract features about the goal context. 



To overcome this challenge, at the performance plateau of the \emph{generalist}, we split the environment into 5 environment variations, each with a different goal (context interval), and initiate 5 \emph{specialists} from the generalist checkpoint to master each of the variations. For each specialist, because the reward distribution is identical across environment resets, the specialist can quickly succeed even though it does not understand the semantic meaning of the context. Then, we use specialists to generate demonstrations. In practice, auxiliary rewards can be generated from the demos using techniques such as Demonstration Augmented Policy Gradient (DAPG, \cite{dapg}) and Generative Adversarial Imitation Learning (GAIL, \cite{gail}). With strong rewards obtained from demonstrations, the generalist is not bothered by the challenges in path-finding and can focus on discriminating goal context.
As a result, it quickly learns (as shown in Fig. \ref{fig:ill-res}) to factor the previously-ignored goal context into making its decisions, thereby overcoming the performance plateau.

Besides that catastrophic ignorance can be cured by our framework, GSL can also deal with the catastrophic forgetting issue. As the generalist learning proceeds, it becomes increasingly hard for its policy/value network to maintain the predictive power for all the environment variations without forgetting. The introduction of specialists can mitigate this issue, because each specialist usually only needs to work well on a smaller subset of the environment variations and these specialized knowledge is transferred and consolidated into a single agent with the help of the collected demonstrations.
As catastrophic forgetting is well-known to the neural network community, we do not provide an illustrative example here.




\algnewcommand{\algorithmicforeach}{\textbf{for each}}
\algdef{SE}[FOR]{ForEach}{EndForEach}[1]
  {\algorithmicforeach\ #1\ \algorithmicdo}
  {\algorithmicend\ \algorithmicforeach}

\begin{algorithm}
  \caption{GSL: Generalist-Specialist Learning}
  \begin{algorithmic}[1]
    \Require (1) Environment $E$ with context space $C$ (2) Number of specialists $N_s$ (3) Number of env. variations for specialist $N_{lenv}$ (4) Number of demonstrations $N_D^g$ from generalist and $N_D^s$ from specialists (5) Performance plateau criteria $\mathcal{H}$
    \State Initialize generalist policy $\pi^g$
    \State Train $\pi^g$ on $E$ until $\mathcal{H} = 1$  \algorithmiccomment{e.g., PPO, SAC}
    \If {$\pi^g$ optimal enough}
        \State Exit \algorithmiccomment{done with GSL}
    \EndIf 
    \State Find the $N_{lenv}$ lowest-performing environment variations from $E$, collectively denoted as $E_{low}$.
    \State Split $E_{low}$ into $N_s$ disjoint environment variations $\{E_i\}$ by splitting the context space $C$.
    \State Obtain $\pi^g_{low}$ by fine-tuning $\pi^g$ on $E_{low}$ \algorithmiccomment{optional} 
    \ForEach {$i=1\cdots N_s$} \algorithmiccomment{in parallel}
        \State Initialize specialist $\pi^s_i=\pi^g$ or $\pi^g_{low}$
        \State Train $\pi^s_i$ on $E_i$ 
        \State Generate $\frac{N_D^s}{N_s}$ demos $\mathcal{T}_i$ with $\pi^s_i$ on $E_i$
    \EndForEach
    \State Generate $N_D^g$ demos $\mathcal{T}_g$ with $\pi^g$ on $E \backslash E_{low}$
    \State Continue training $\pi_g$ on $E$ with auxiliary rewards induced from $\{\mathcal{T}^D_i\} \cup \mathcal{T}_g$ (via DAPG, GAIL, etc.)
  \end{algorithmic}
  \label{alg:gsl}
\end{algorithm}

%% file: sections/method.tex
\section{Generalist-Specialist Learning}

Motivated by our previous example, we now introduce our GSL framework. At a high-level, the framework is a ``meta-algorithm" that integrates a reinforcement learning algorithm and a learning-from-demonstration algorithm as building blocks and produce a more powerful reinforcement learning algorithm. While there exists works with similar spirit, we identify several design choices that are crucial to the success but were not revealed in the literature. We will first describe the basic framework, and then introduce our solutions to the key design choices which lead to improved sample complexity in environments that are too difficult for the building block reinforcement learning algorithm due to the catastrophic forgetting and ignorance issues. 

\subsection{The Meta-Algorithm Framework} 
We first initialize a generalist policy $\pi^g$ and train the model over all variations of the environment using actor-critic algorithms such as PPO~\citep{PPO} and SAC~\citep{sac}. When the performance plateau criteria $\mathcal{H}$ (introduced later) is satisfied, we stop the training of $\pi^g$. This could occur either when $\pi^g$ reaches optimal performance (in which case we are done), or when the performance is still sub-optimal. If the performance is sub-optimal, we then split all environment variations into small subsets, and launch a population of specialists, each initiated from the checkpoint of generalist, to master each subset of variations. We finally obtain demonstrations from the specialists, and resume generalist training with auxiliary rewards created by these demonstrations. The basic framework is outlined in Algorithm \ref{alg:gsl}.


\subsection{When and How to Train Specialists} 
\label{sec:when_how_train_specialists}
While there exist attempts on using the divide-and-conquer strategy to solve tasks in diverse environments, a systematic study over the timing to start specialist training is missing. The default choice in the literature \citep{teh2017distral, ghosh2017divide} starts specialist training from the very beginning and periodically distills the specialists into the generalist. However, as in Fig. \ref{fig:specialist_timing}, we observe that training the specialists before the generalist's performance plateaus does not take full advantage of generalist's fast learning during the early stages, and therefore results in less optimal sample complexity. Consequently, we start specialist training only after the generalist's performance plateaus.

\textbf{Performance plateau criteria for generalists.} We introduce a binary criteria $\mathcal{H}$ to decide when the performance of generalist plateaus. 
In our implementation, we design a simple but effective criteria based on the change of average return, which works well in our benchmarks. Given returns from $M$ epochs $\{R_1, \dots, R_M\}$, we first apply a 1D Gaussian filter with kernel size 400 to smooth the data. Then, $\mathcal{H}(t) = \mathbbm{1}(R_t + \epsilon \ge R_{t'},\ \forall\ t' \in\ \{t+1, t+2 ..., t+W\}).$
Intuitively, the criteria is satisfied if the smoothed return at a certain epoch is approximately higher than (more than a margin $\epsilon$) all smoothed returns in the future $W$ epochs. 

\textbf{Assigning environment variations to specialists.} When we assign sets of environment variations to specialists, we hope that each specialist can master their assigned variations, yet we also hope that the number of specialists $N_s$ is not too large if the number of all training environment variations is already large (in which case we need a large amount computational resource to train the specialists in parallel). Empirically, we observe that the generalist can solve some environment variations reasonably well, yet performs poorly on others. Therefore, we launch specialist training only on the $N_{lenv}$ lowest performing environment variations. 

We empirically find that an optional step (line 8 of Alg. \ref{alg:gsl}), specifically fine-tuning $\pi^g$ on the $N_{lenv}$ lowest performing environment variations before training the specialists, can help to improve sample efficiency as it gives the specialists a better starting point.
Specifically, we fine-tune $\pi^g$ for $200$ epochs, or $200*16384$ samples (a very small number compared to the 100M total budget) and find this step helpful for tasks in Procgen if PPO is used as the backbone RL algorithm.

After we assign the variations to specialists, we start to train these specialists in parallel. We assume that a specialist can always solve the environment with a few variations. For example, for most tasks in Procgen which contain 1024 procedurally generated levels for training, we find $N_{lenv}=300$ and $N_{s}=75$ good enough (i.e. each specialist is trained on $300/75=4$ variations). Therefore, we can train the specialists until they accomplish their assigned variations. In practice, we set a fixed number of samples $N_{sample}$ for training each specialist.

\subsection{Generalist Training Guided by Specialist Demos} 
\label{sec:specialist2generalist}
After training each specialist to master a small set of environment variations, we still need the common generalist to consolidate specialist experiences and master \emph{all} training environment variations. In our proposed framework, we first collect the demonstration set $\{\mathcal{T}^{D}_i\}$ using the specialists on their respective training environment variations (we only collect trajectories whose rewards are greater than a threshold $\tau$). 
Specifically, we use the best performing model checkpoint stored of each specialist to generate the demonstration set.
To ensure training stability, we also collect demos $\mathcal{T}_g$ for the remaining training variations using the generalist.
We then resume generalist training using a \emph{learning-from-demonstrations algorithm} by combining the environment reward and the auxiliary rewards induced from these demonstrations. To train a generalist in this process, we can adopt many approaches, such as Behavior Cloning (BC), Demonstration-Augmented Policy Gradient (DAPG, \cite{dapg}), and Generative-Adversarial Imitation Learning (GAIL, \cite{gail}). 

It is a key design factor to choose this learning-from-demonstrations algorithm. The crucial challenge comes from the inconsistency of specialist behaviors in similar states. To our knowledge, previous works of divide-and-conquer RL uses BC to distill from specialists in an offline manner; however, even if different environment variations share the same reward structure and various regularization techniques can be added to the specialist training process, we find it not scalable to the number of specialists and it fails to work well in diverse environments such as Procgen or ManiSkill in our paper.
Moreover, pure offline methods such as BC usually achieve inferior performance since they are limited to a fixed set of demonstrations.



With the demos collected from specialists, we find online learning-from-demonstration methods such as DAPG and GAIL to be quite effective. 
For DAPG and GAIL, besides utilizing all the collected demos $\{\mathcal{T}^{D}_i\} \cup \mathcal{T}_g$, we also let the generalist interact with the environment to obtain online samples. From here, we use $\rho_{D}$ and $\rho_{\pi}$ to denote a batch of transitions sampled from the demonstrations and from the environment, respectively. While DAPG and GAIL in principle can be adapted to any RL algorithms (PPO, PPG, SAC, etc.), we evaluate GSL on challenging benchmarks using their corresponding strong baseline RL algorithms (PPO/PPG on Procgen, PPO on Meta-World, and SAC on ManiSkill).
Below, we derive formula to illustrate how we adapt DAPG and GAIL in our experiments.

We modify DAPG for DAPG + PPO. We first calculate the advantage value $A(s,a)$ for $(s,a) \sim \rho_{\pi}$ using GAE \citep{GAE}. Then, in each PPO epoch, we compute the maximum advantage denoted as $\hat{A}$. We obtain the overall policy loss (value loss omitted here):



\begin{gather*}
\mathcal{L}_{\rho}^{C L I P}(\theta)= - {\mathbb{E}}_{(s,a) \sim \rho} \\ \scriptstyle  \left[\min \left(\frac{\pi_{\theta}(a|s)}{\pi_{\theta_{o l d}}(a|s)}A(s,a),  \operatorname{clip}(r_{t}(\theta), 1-\epsilon, 1+\epsilon) A(s,a)\right)\right] \\
\mathcal{L}_{\rho}^{1}(\theta) = - {\mathbb{E}}_{(s,a) \sim \rho} [\pi_{\theta}(a|s)] \\
\mathcal{L}_{DAPG+PPO}(\theta) = \mathcal{L}_{\rho_{\pi}}^{C L I P}(\theta) + \omega\cdot \hat{A}\cdot \mathcal{L}_{\rho_{D}}^{1}(\theta)  \\
\end{gather*}

We find that a smoothed loss $\mathcal{L}^1(\cdot)$ here significantly improves training stability for some tasks, compared to the cross entropy (style) loss used in the original DAPG paper.
We set $\omega=0.5$ in all our experiments and find that decreasing it over time (as in the original DAPG paper) can lead to worse performance.

For GAIL + SAC, we train a discriminator to determine whether a transition comes from policy or comes from demonstration. We obtain the following losses for policy $\pi_{\phi}$, discriminator $D_{\psi}$, and Q-function $Q_{\theta}$:

\begin{gather*}
\mathcal{L}_{GAIL+SAC} = \mathcal{L}_{\pi}(\phi) + \mathcal{L}_{D}(\psi) + \mathcal{L}_{Q}(\theta) \\
\scriptstyle \mathcal{L}_{\pi}(\phi) = -\mathbb{E}_{s_t \sim \rho_{\pi}}\left[\mathbb{E}_{a_{t} \sim \pi_{\phi}}\left[\alpha \log (\pi_{\phi}(a_{t} | s_{t}))-Q_{\theta}(s_{t}, a_{t})\right]\right] \\
\scriptstyle \mathcal{L}_{D}(\psi) = \mathbb{E}_{\rho_{\pi}}[\log (D_{\psi}(s_t, a_t))]+\mathbb{E}_{\rho_{D}}[\log (1-D_{\psi}(s_t, a_t))] \\
\scriptstyle \mathcal{L}_{Q}(\theta) = \mathbb{E}_{\rho_{\pi} \cup \rho_{D}}[(Q_{\theta}(s_{t}, a_{t})-(\tilde{r}(s_{t}, a_{t})+\gamma V_{\bar{\theta}}(s_{t+1})))^{2}] \\
\tilde{r}(s_{t}, a_{t}) = \beta r(s_{t}, a_{t}) + (1-\beta) \log (D_{\psi}(s_t, a_t))
\end{gather*}

here $\alpha$ is the temperature in SAC; $\beta$ is the hyper-parameter thlat interpolates between the environment reward and the reward from the discriminator; moreover,
$V_{\bar{\theta}}(s_{t}) = \mathbb{E}_{a_t \sim \pi_{\phi}(\cdot | s_{t})} \left[ Q_{\theta}(s_t, a_t) - \alpha \log (\pi_{\phi}(a_{t} | s_{t})) \right]$ is target value.
More implementation details are in Appendix.

PPG \cite{cobbe2021phasic} is a recently proposed RL algorithm that equips PPO with an auxiliary phase for learning the value function.
It achieves state-of-the-art training performance on Procgen.
We therefore also evaluate the PPG + DAPG combination on Procgen, with essentially the same adaptation as for PPO + DAPG.
Our experiments illustrate the generality of our framework as a meta-algorithm.

%% file: sections/experiments.tex
\begin{figure*}[t]
\centering

\begin{subfigure}[b]{\linewidth}
    \begin{subfigure}[b]{\linewidth}
        \centering
        \includegraphics[width=0.24\linewidth]{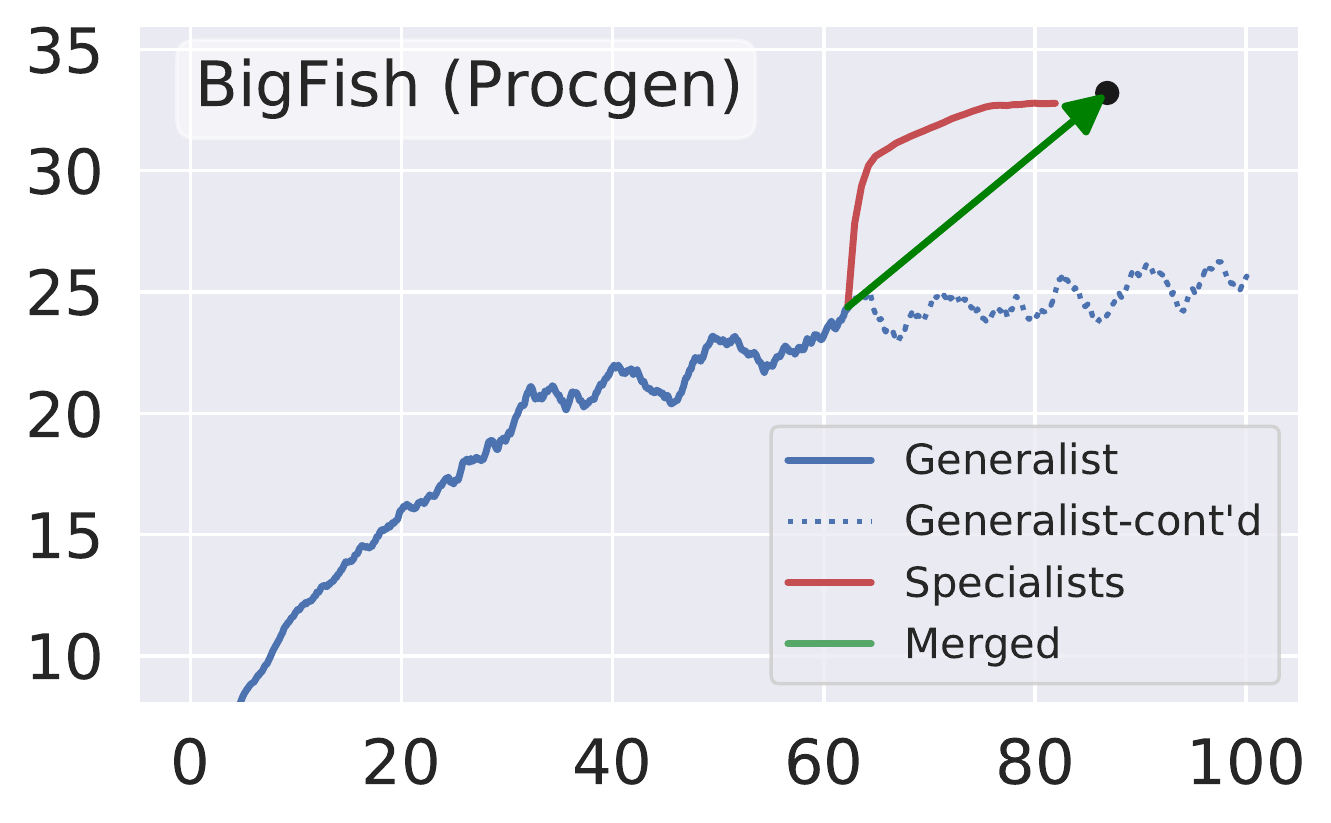}
        \includegraphics[width=0.24\linewidth]{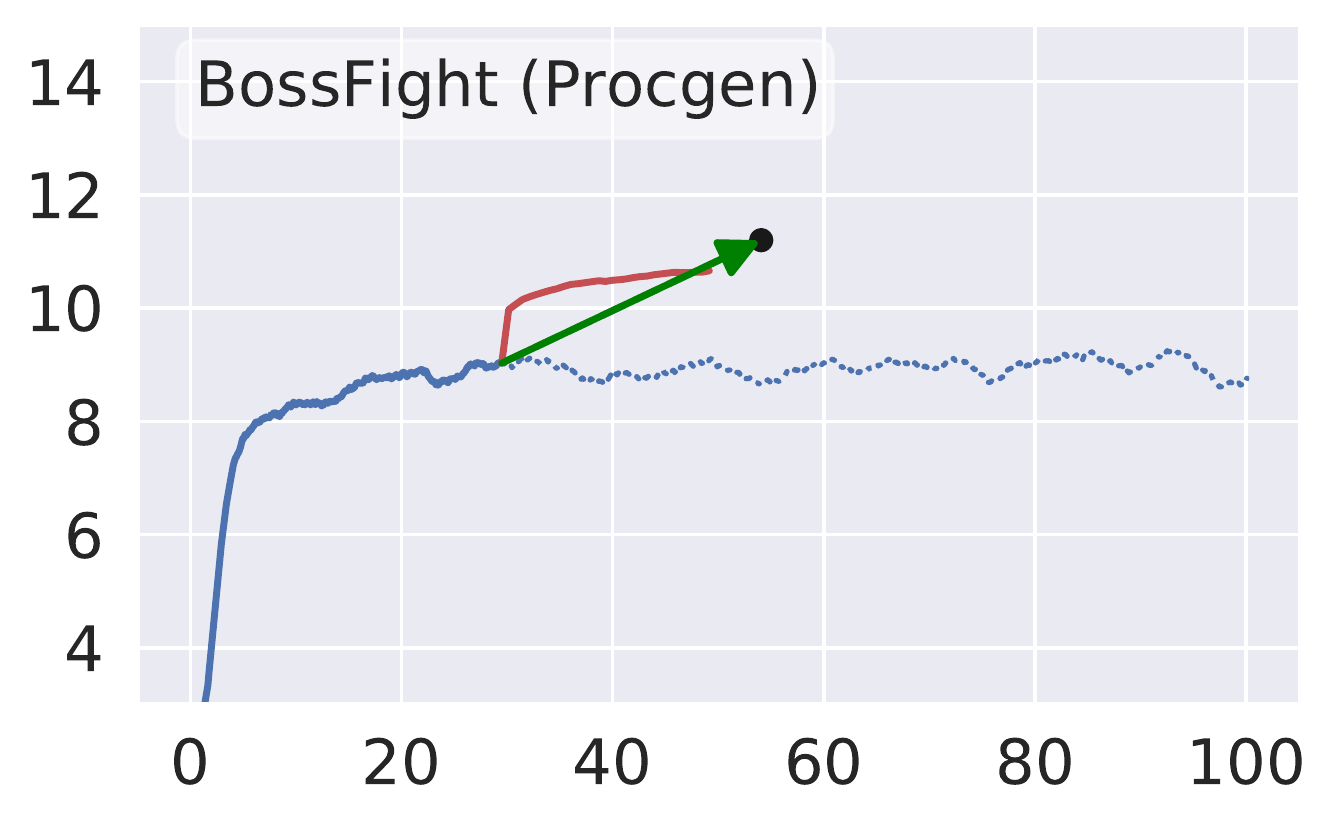}
        \includegraphics[width=0.24\linewidth]{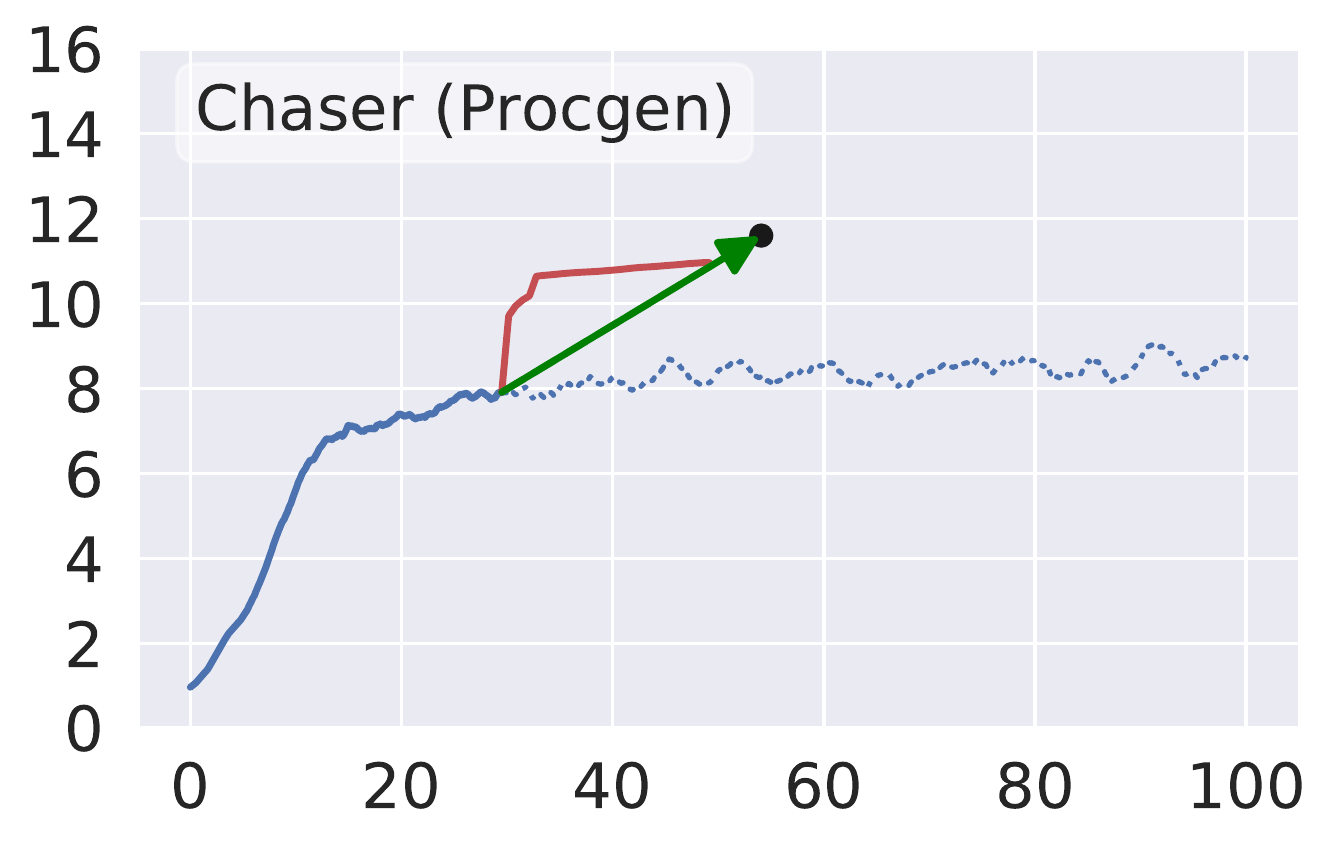}
        \includegraphics[width=.24\linewidth]{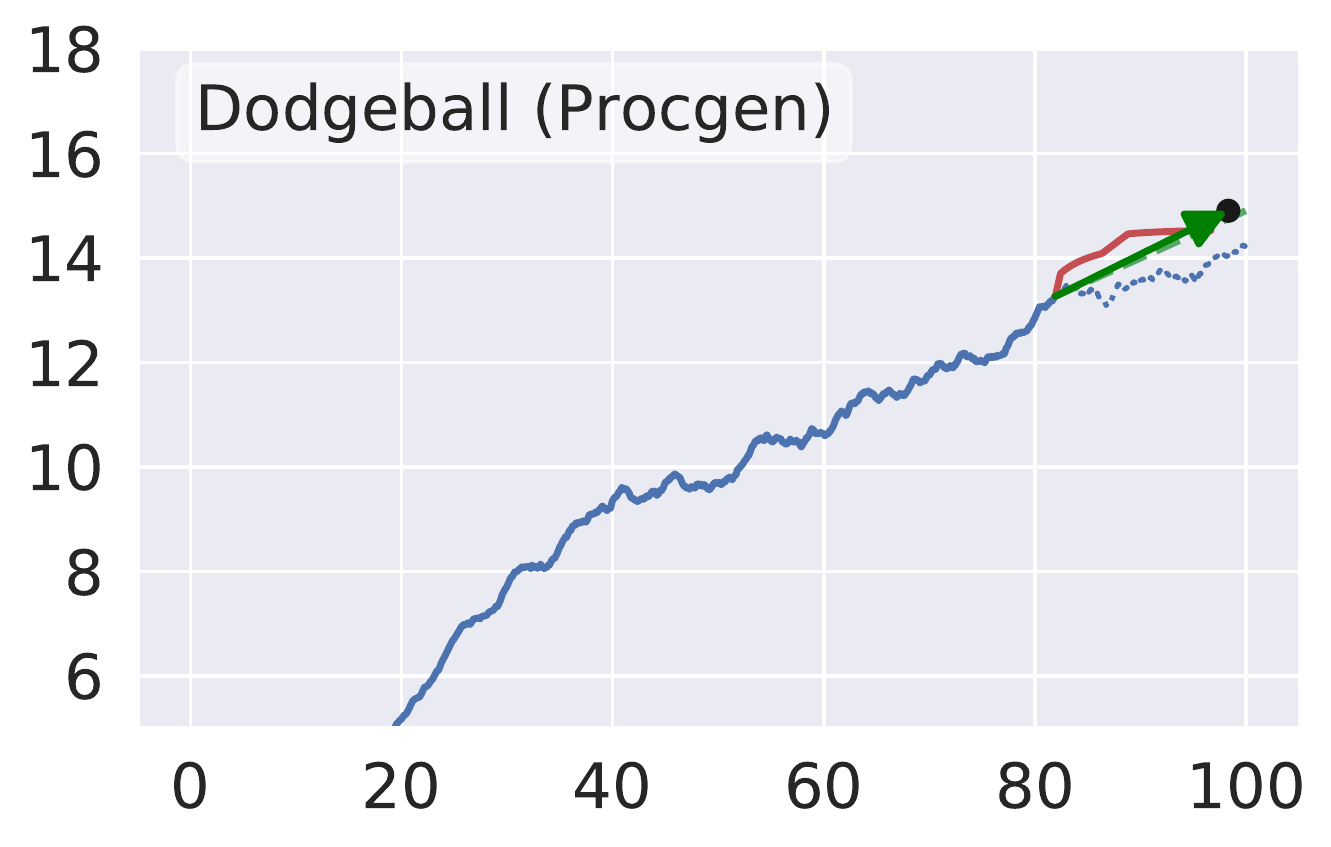}
    \end{subfigure}
    \begin{subfigure}[b]{\linewidth}
        \centering
        \includegraphics[width=0.24\linewidth]{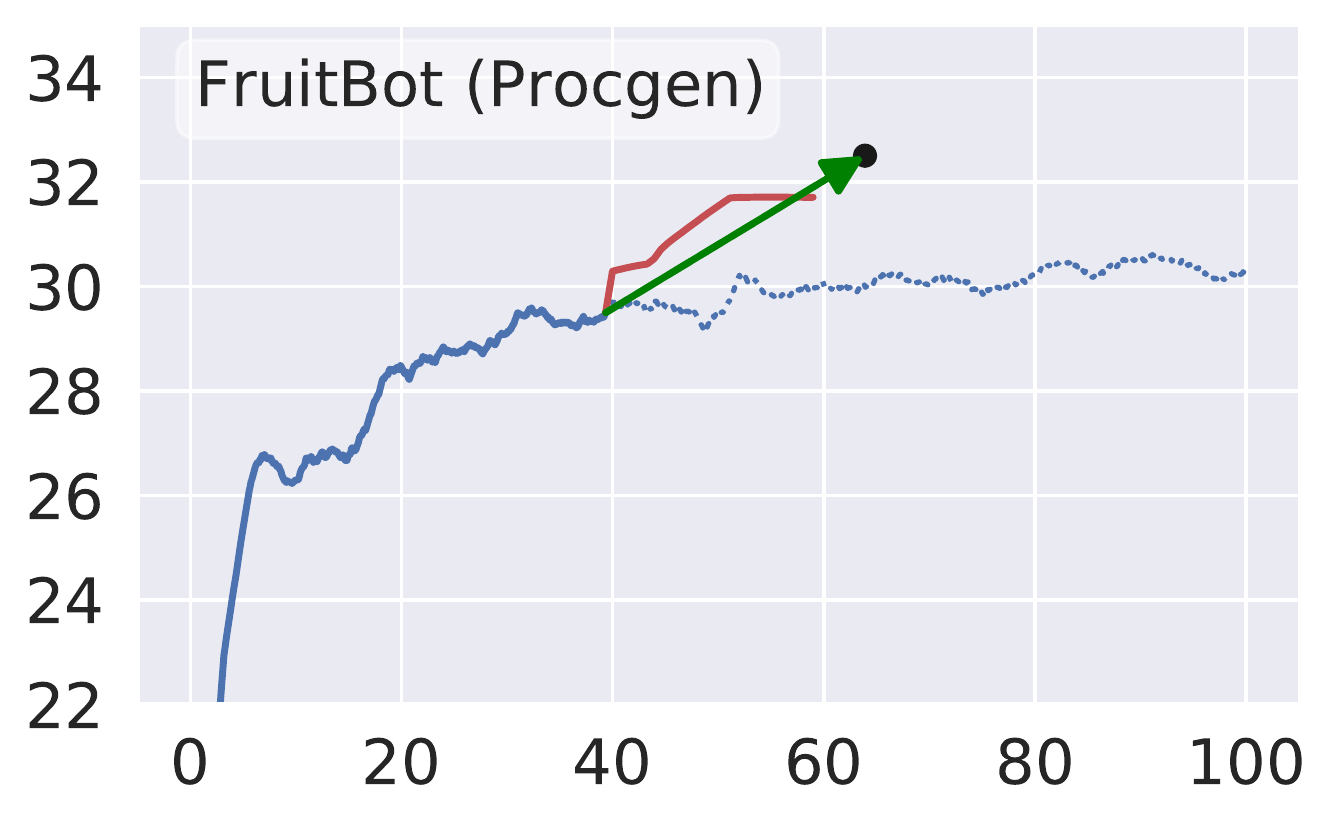}
        \includegraphics[width=0.24\linewidth]{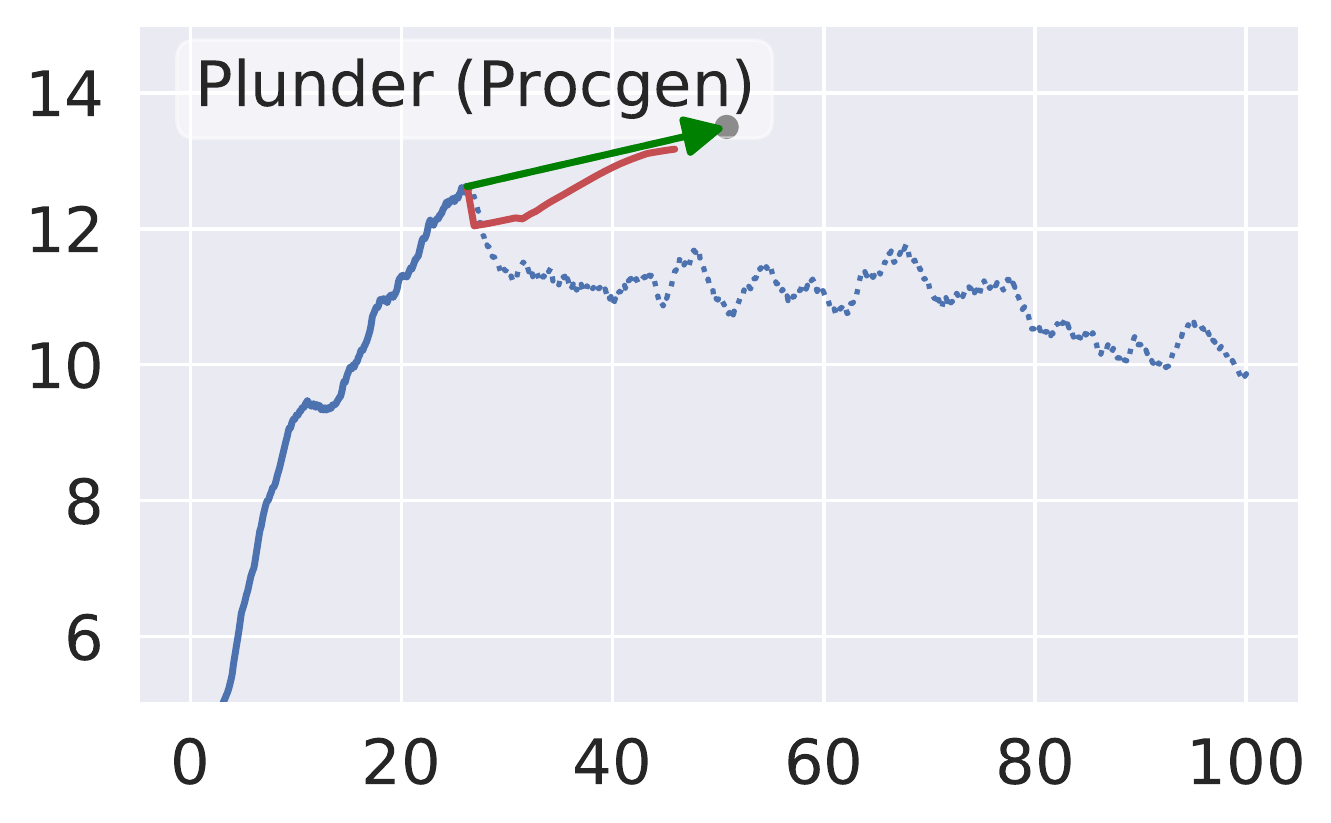}
        \includegraphics[width=0.24\linewidth]{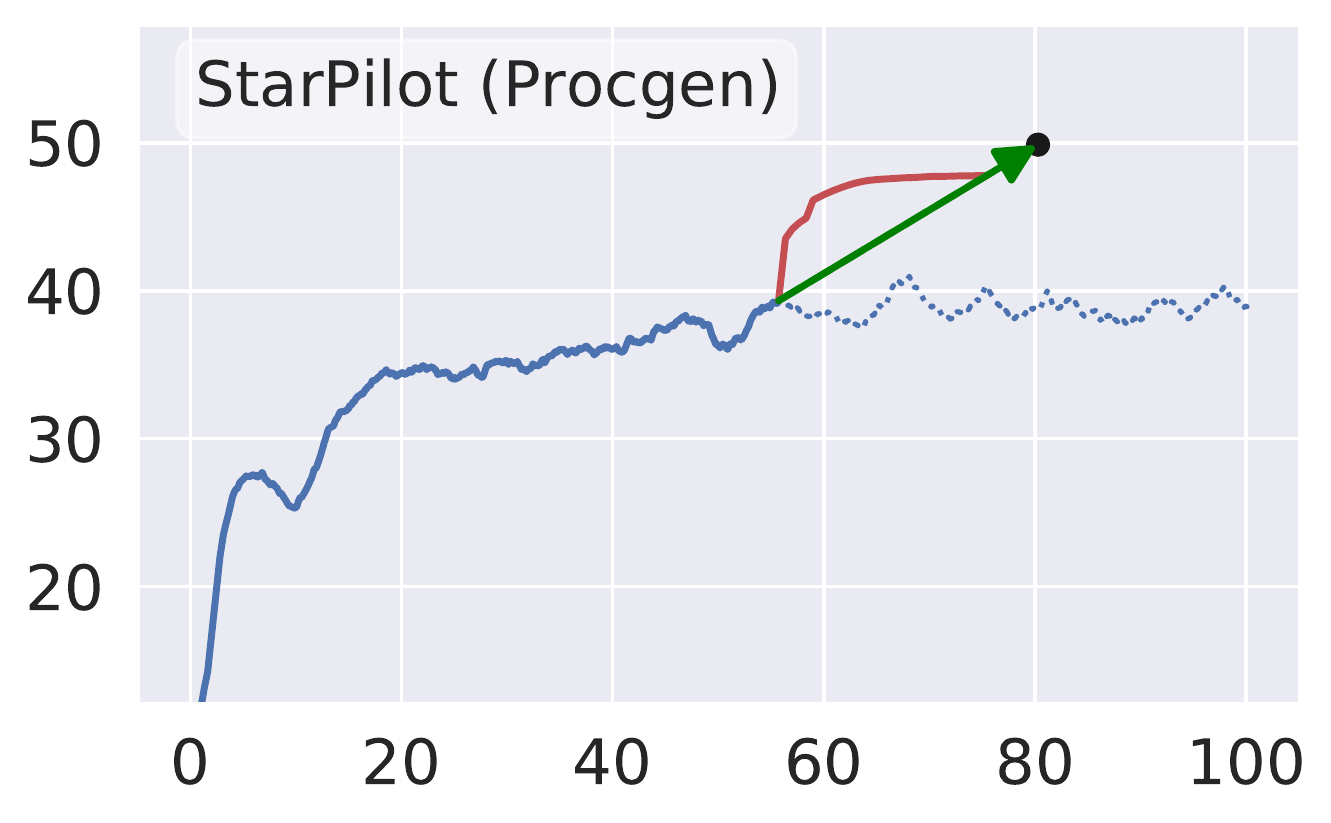}
        \includegraphics[width=0.24\linewidth]{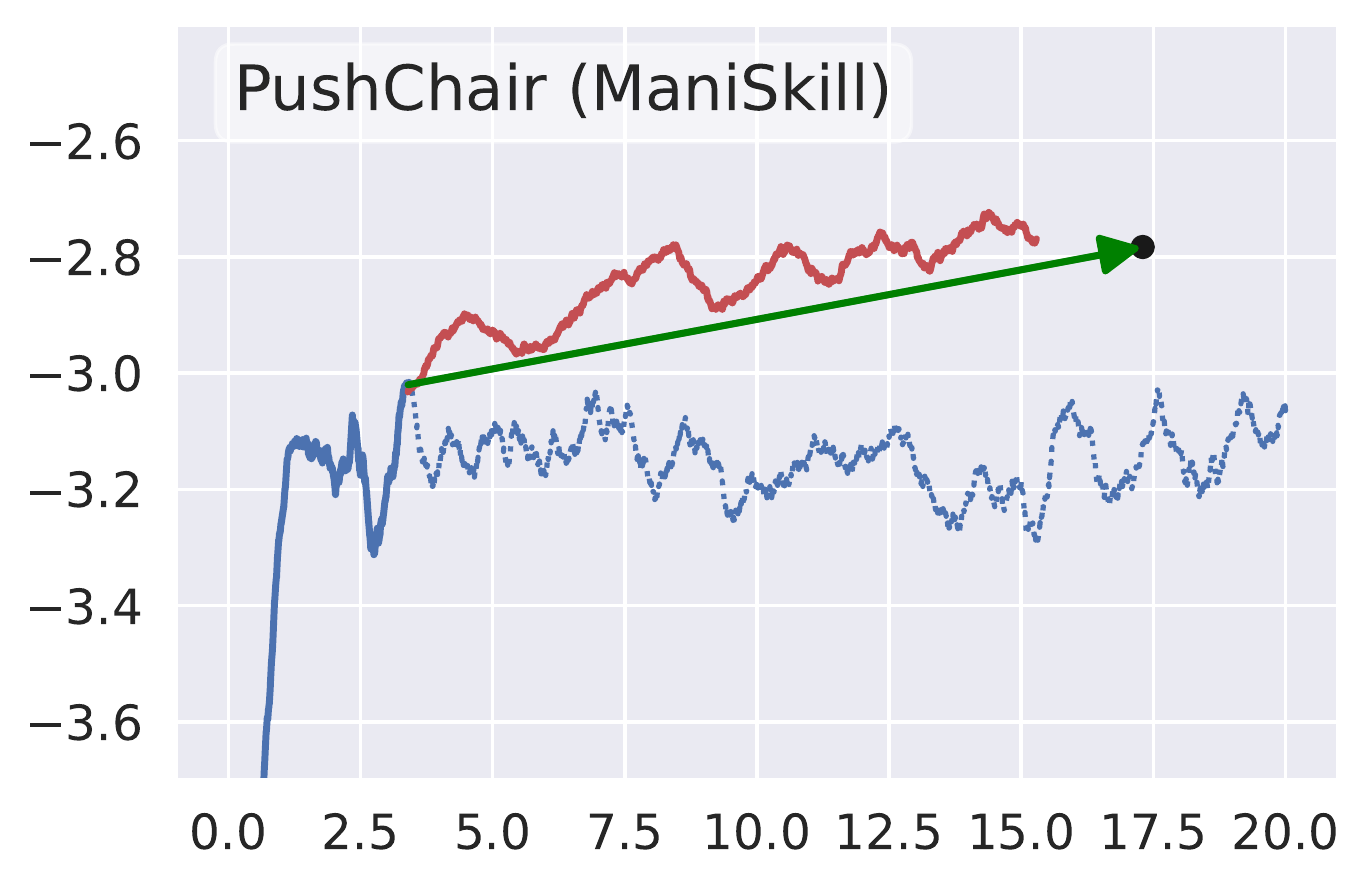}
    \end{subfigure}
\end{subfigure}
\vspace{-1mm}

\caption{GSL significantly improves baselines (in blue) on Procgen (PPO) and ManiSkill (SAC) for their most challenging environments, respectively. X-axis is the overall training samples (M), and y-axis is the return. The unit for the y-axis in PushChair is $1000$. The dashed blue line indicates what happens if we choose to continuously train the generalist. For clarity, we only show the starting and ending points for the step of converting specialist experiences into the generalist (green arrow) and only plot one run (See Appendix for more details including training curves across all runs with mean and std.). We report the numerical results in Tab. \ref{tab:main_procgen_train}.}
\label{fig:overall}
\end{figure*}

\begin{figure*}[t]
\centering

\begin{subfigure}[b]{\linewidth}
    \begin{subfigure}[b]{\linewidth}
        \centering
        \includegraphics[width=0.24\linewidth]{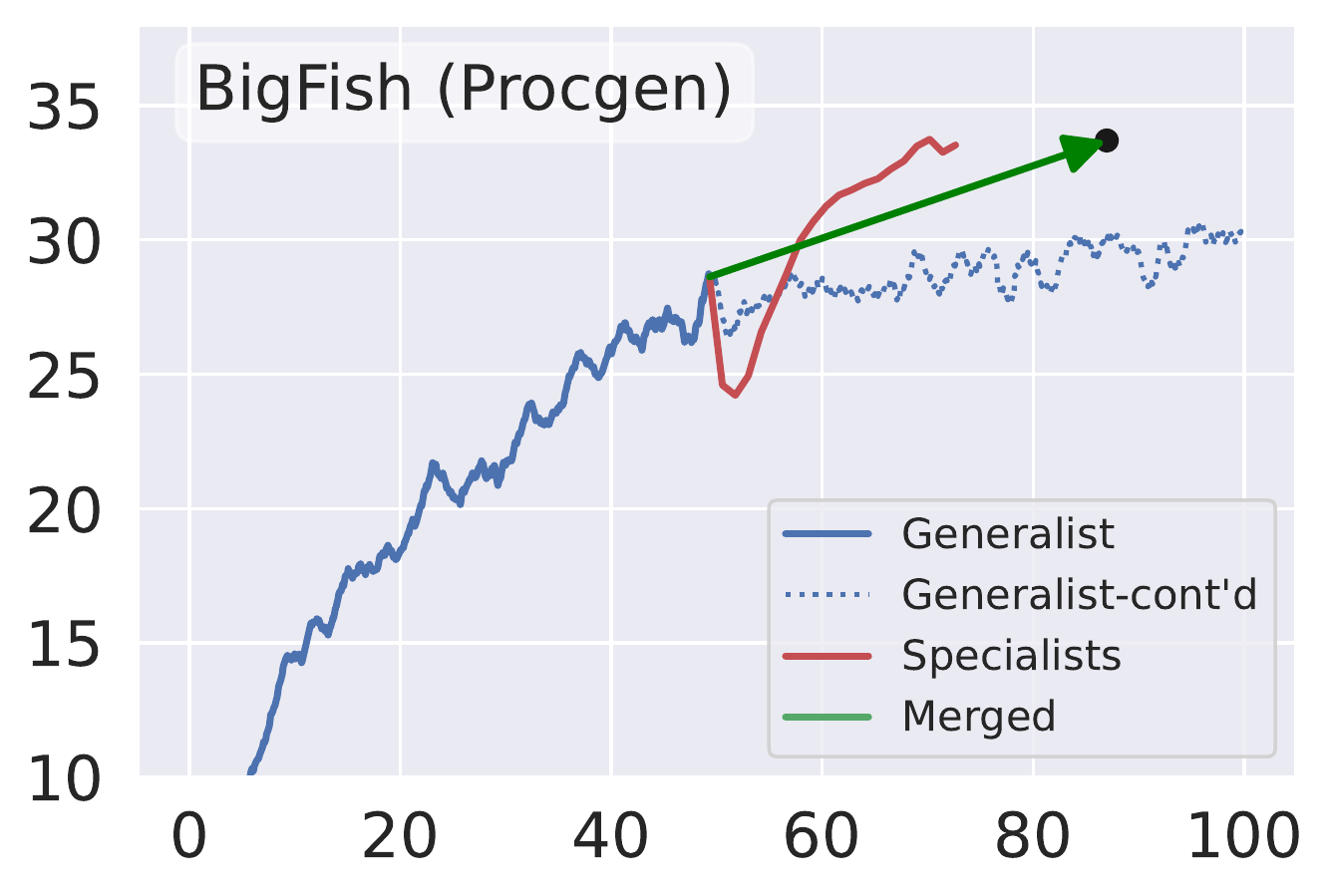}
        \includegraphics[width=0.24\linewidth]{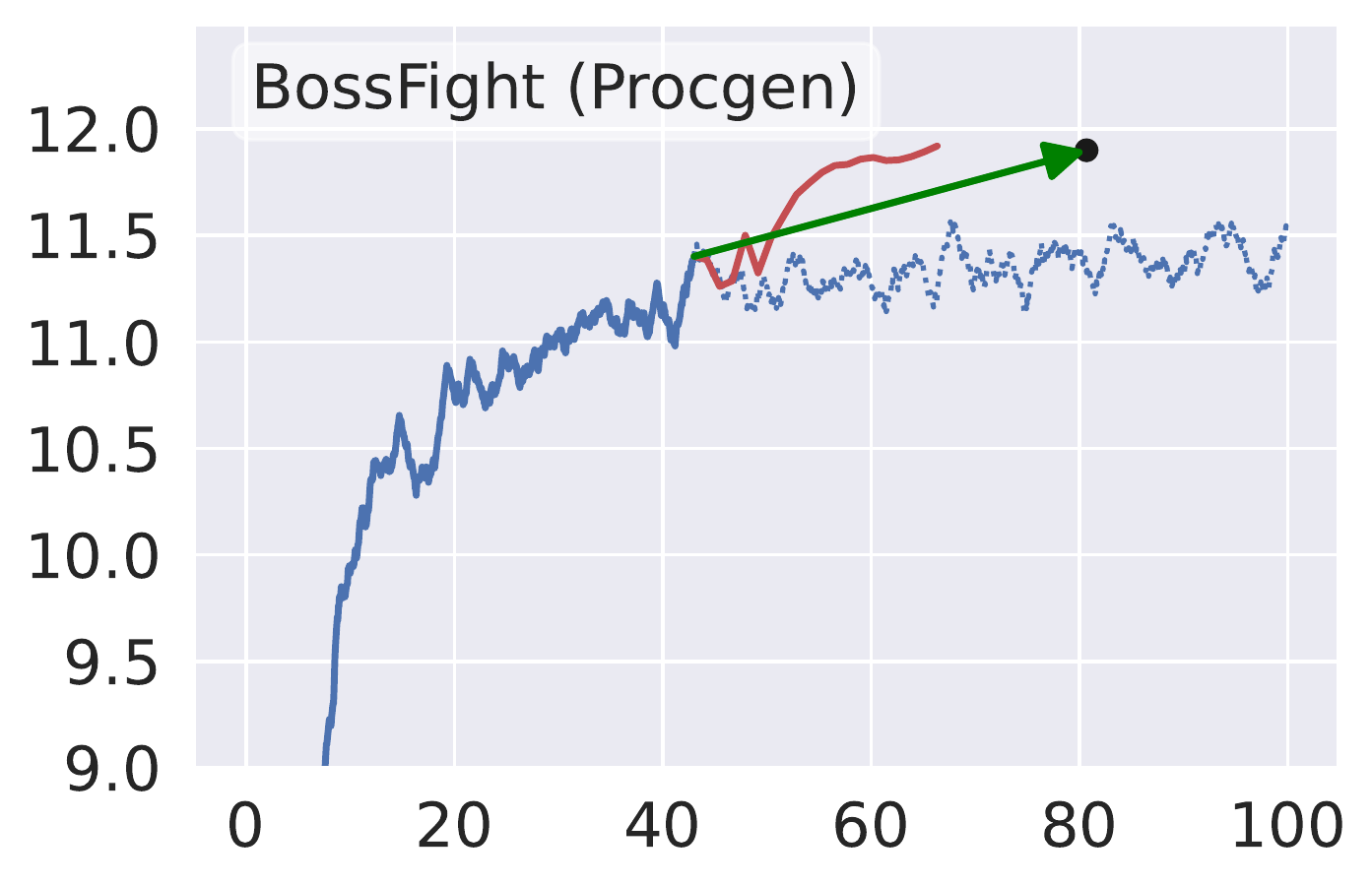}
        \includegraphics[width=0.24\linewidth]{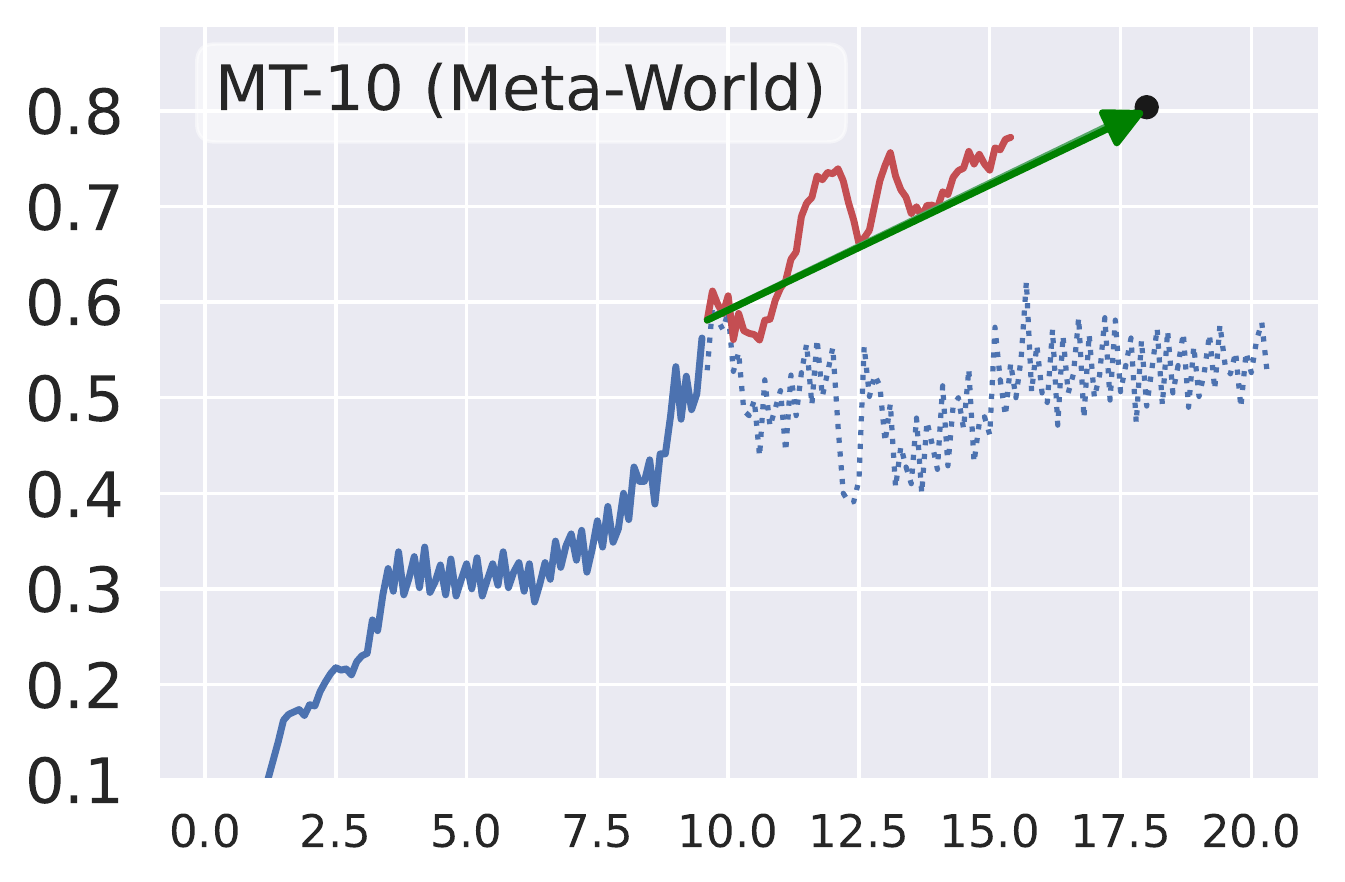}
        \includegraphics[width=0.24\linewidth]{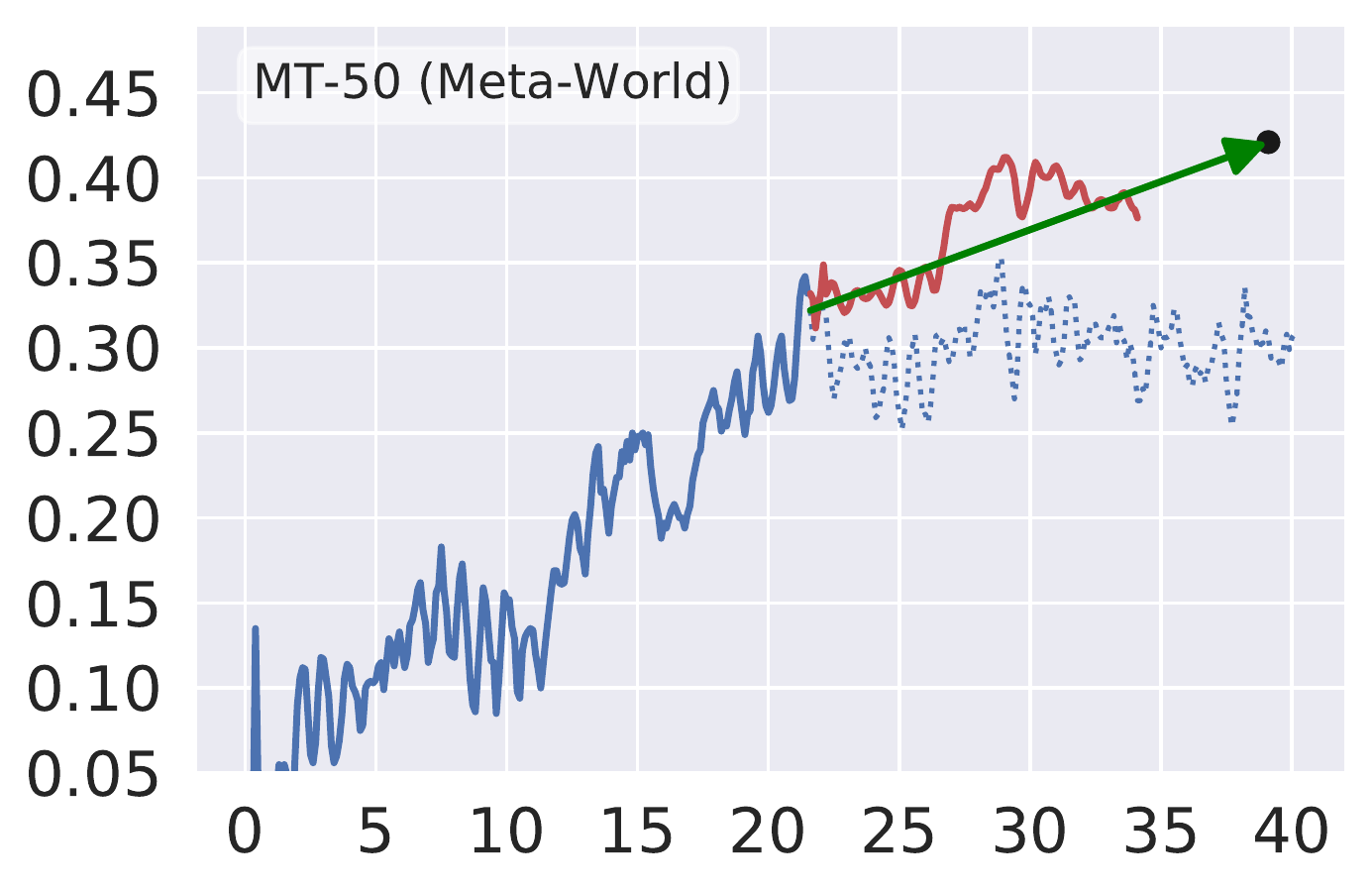}
    \end{subfigure}
\end{subfigure}
\vspace{-1mm}

\caption{We also evaluate GSL on MT-10 \& MT-50 (PPO) and on the first two challenging tasks of Procgen (with PPG as the baseline, which already achieves close-to-optimal performance). Again, GSL consistently improves the baselines. The unit for the x-axis is million. Similar to Fig. \ref{fig:overall}, we only plot one run out of 5 for clarity. See Tab. \ref{tab:main_procgen_train} \& \ref{tab:extra} for numerical results and the Appendix for more details.}
\label{fig:extra}
\end{figure*}

\begin{table*}[h!]
\centering
\scalebox{0.83}{
\begin{tabular}{c|ccccccc} 
 \hline
 & BigFish & BossFight & Chaser & Dodgeball & FruitBot & Plunder & StarPilot \\
 \hline
 PPO-Train & 24.6$\pm$0.7 & 8.6$\pm$0.2 & 8.5$\pm$0.3 & 13.7$\pm$0.3 & 30.1$\pm$0.6 & 10.5$\pm$0.8 & 39.4$\pm$1.4 \\
 GSL-Train & \textbf{31.1$\pm$0.8} & \textbf{11.3$\pm$0.2} & \textbf{11.5$\pm$0.3} & \textbf{15.5$\pm$0.2} & \textbf{31.9$\pm$0.3} & \textbf{13.4$\pm$0.4} & \textbf{49.5$\pm$0.4} \\
  \hline
 PPO-Test & 24.3$\pm$1.1 & 8.6$\pm$0.3 & 7.9$\pm$0.4 & 12.7$\pm$0.3 & 29.1$\pm$0.5 & 9.7$\pm$0.5 & 38.0$\pm$0.9 \\
 GSL-Test & \textbf{30.0 $\pm$0.5} & \textbf{10.4 $\pm$0.2} & \textbf{10.9$\pm$0.2} & \textbf{14.1$\pm$0.3} & \textbf{30.5$\pm$0.4} & \textbf{13.1$\pm$0.3} & \textbf{48.7$\pm$0.5} \\
 \hline
\end{tabular}
}
\scalebox{0.83}{
\begin{tabular}{c|cc} 
 \hline
 & BigFish & BossFight  \\
 \hline
 PPG-Train & 29.4$\pm$1.1 & 11.3$\pm$0.2\\
 GSL-Train & \textbf{33.5$\pm$1.3} & \textbf{11.9$\pm$0.2} \\
  \hline
 PPG-Test & 28.0$\pm$0.9 & 11.1$\pm$0.2\\
 GSL-Test & \textbf{30.9 $\pm$0.8} & \textbf{ 11.6$\pm$0.2}\\
 \hline
\end{tabular}
}
\caption{Our GSL framework outperforms PPO and PPG baselines on both training and generalization. Models are trained over 1024 levels for each environment and tested over 1000 unseen levels. Results are averaged over 5 runs with std. from raw episode rewards.}
\label{tab:main_procgen_train}
\end{table*}

\subsection{Benchmarks}
We evaluate our Generalist-Specialist Learning (GSL) framework on three challenging benchmarks: Procgen~\citep{procgen}, Meta-World~\citep{yu2020meta} and SAPIEN Manipulation Skill Benchmark (ManiSkill Benchmark~\citep{mu2021maniskill}). 

Procgen is a set of 16 vision-based game environments, where each environment leverages seed-based procedural generation to generate highly-diverse levels. All Procgen environments use 15-dimensional discrete action space and produce $(64,64,3)$ RGB observation space. In our experiments, we use 1024 levels for training, which is different from the original 200-training-level setup (as we try to evaluate how well GSL scales). We select the $7$ most challenging environments under our setting based on the normalized score obtained by training the baseline PPO algorithm (PPO can achieve a close to perfect score on the remaining environments given 100M total samples). These environments are BigFish, BossFight, Chaser, Dodgeball, FruitBot, Plunder, and StarPilot. A subset of environments are illustrated in Fig. \ref{fig:prcgen-env}. This benchmark leverages procedural generation and is also suitable for  evaluating the generalization performance of our framework. We adopt the IMPALA CNN model \cite{espeholt2018impala} as the network backbone.

Meta-World is a large-scale manipulation benchmark for meta RL and multi-task RL featuring 50 distinct object manipulation skills (see samples in Fig. \ref{fig:metaworld-env}).
We choose multi-task RL as our testbed, including MT-10 and MT-50 (with 10 and 50 skills to learn, respectively), which only evaluate the agents' performance on the training environments.
The state space for this benchmark is high-dimensional and continuous, representing coordinates \& orientations of target objects and parameters for robot arms as well as encodings for the skill ID.
We use the V2 version of the benchmark. See Appendix for the network details.

ManiSkill is a recently proposed benchmark suite designated for learning generalizable low-level physical manipulation skills on 3D objects. The diverse topological and geometric variations of objects, along with randomized positions and physical parameters, lead to challenging policy optimization. Since realistic physical simulation and point cloud rendering processes (see Fig. \ref{fig:maniskill-env}) are very expensive for ManiSkill environments (empirically we observe that the speed is at 30 environment steps per second), we only evaluate our framework on the PushChair task. In this task, an agent needs to move a chair towards a goal through dual-arm coordination (see Fig. \ref{fig:maniskill-env}). The environment has a 22-dimensional continuous joint action space. Each observation consists of a panoramic 3D point cloud captured from robot cameras and a 68-dimensional robot state (which includes proprioceptive information such as robot position and end-effector pose). In our experiments, we use a smaller scale of the PushChair environment that consists of 8 different chairs, where we already find our framework significantly improves the baseline. We also transform all world-based coordinates in the observation space into robot-centric coordinates. Different from the original environment setup, we add an indicator of whether the robot joints experience force feedback due to contact with objects, and we do not reset the environment until the time limit is reached. We adopt the PointNet + Transformer over object segmentations model in the original baseline as our network backbone.


\subsection{Results}
\label{sec:results}
We first train a baseline PPO model on Procgen and MT-10 and MT-50 from Meta-World, along with a baseline SAC model on ManiSkill (full details in Appendix \ref{app:hyper-all}). We then compare these baselines with our GSL framework where PPO+DAPG is trained on Procgen and Meta-World and SAC+GAIL is trained on ManiSkill. We demonstrate the results in Fig.~\ref{fig:overall} \& \ref{fig:extra} and Tab.~\ref{tab:main_procgen_train} \& \ref{tab:extra}. 
We also perform experiments with PPG+DAPG on the first two tasks of Procgen to further verify the generality of GSL (See Tab. \ref{tab:main_procgen_train}).
Since our GSL framework involves online interactions from specialists, \textit{we perform necessary scaling to reflect the actual total sample complexity of our framework}.

\begin{table}[h!]
\centering
\scalebox{0.76}{
\begin{tabular}{c|cc} 
 \hline
 & MT-10 (\%) & MT-50 (\%)  \\
 \hline
 PPO-Train & 58.4$\pm$10.1 & 31.1$\pm$4.5 \\
 GSL-Train & \textbf{77.5$\pm$2.9} & \textbf{43.5$\pm$2.2} \\
  \hline
\end{tabular}
}
\scalebox{0.76}{
\begin{tabular}{c|c} 
 \hline
 & PushChair (k) \\
 \hline
 SAC-Train & -2.97$\pm$2.7 \\
 GSL-Train & \textbf{-2.78$\pm$2.3}  \\
  \hline
\end{tabular}
}
\vspace{-1mm}
\caption{GSL boosts training efficiency on MT-10/MT-50  and PushChair. Results are averaged over 5 and 3 runs, respectively.}
\label{tab:extra}
\end{table}

We observe that the generalist learns fast at the beginning, yet its performance plateaus at a sub-optimal level. However, as soon as we launch specialist training, they quickly master their assigned variations. The strong rewards obtained from specialists' demonstrations efficiently and effectively lift the generalist out of performance plateaus, resulting in significant improvement over baselines. These observations also corroborate those in Fig. \ref{fig:ill-res} of our illustrative example. 

We compare the final average performance of GSL with PPO/PPG baselines on (1) the 1024 training levels and also (2) the 1000 hold-out test levels of Procgen. We observe that our framework not only improves over the baseline on the training environment variations, but also on unseen environment variations, demonstrating our framework's effectiveness for obtaining generalizable policies.

%% file: sections/ablations.tex
\section{Ablation Studies}
\label{sec:ablations}

In this section, we perform several ablation studies to justify our design. In ablation experiments, we use $256$ different levels for all Procgens experiments, as opposed to $1024$ different levels in the main result section (Sec.~\ref{sec:results}), since obtaining results on fewer levels is faster.


\subsection{Influence of Environment Variations on Training Efficiency and Effectiveness}
\begin{figure}[h]
\centering
\begin{subfigure}[]{0.455\linewidth}
    \centering
    \includegraphics[width=\linewidth]{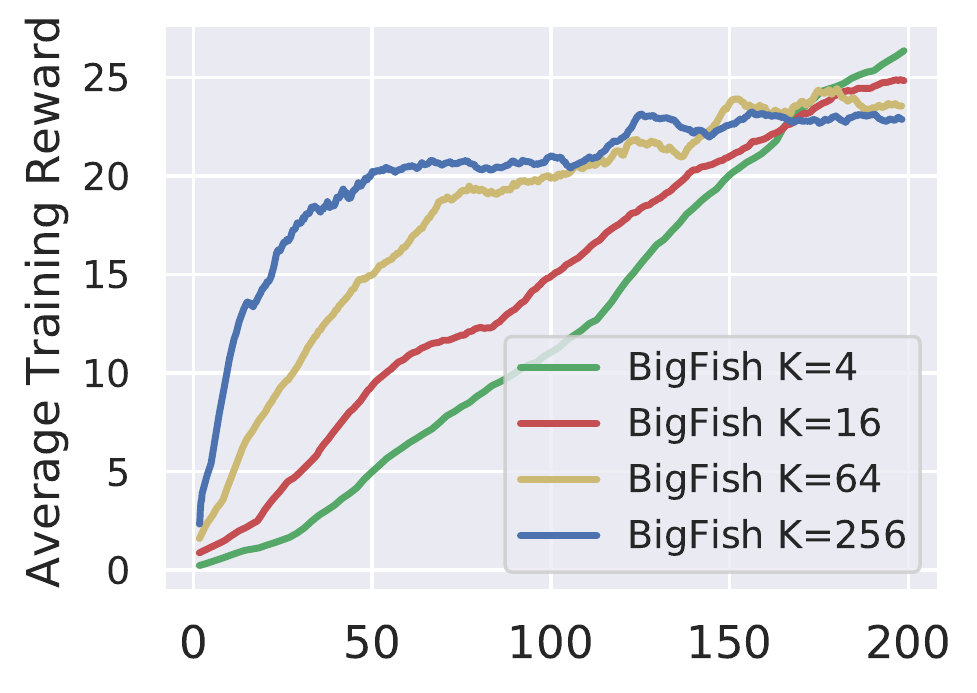}
\end{subfigure}
\begin{subfigure}[]{0.435\linewidth}
    \centering
    \includegraphics[width=\linewidth]{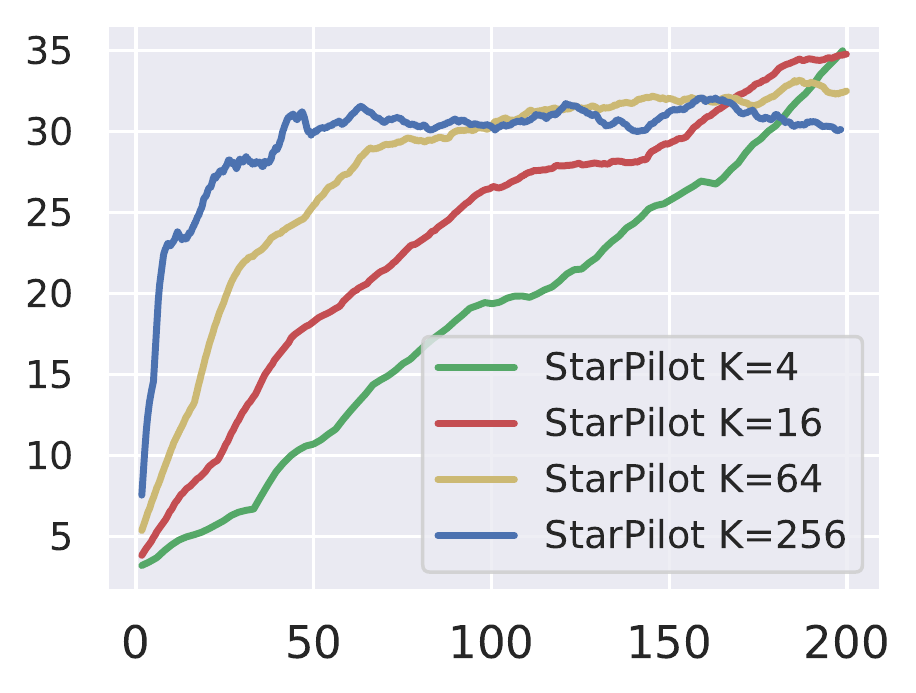}
\end{subfigure}
\vspace{-0.3em}
\caption{The average training returns on BigFish and StarPilot over $256$ levels for $N_s=256/K$ specialists trained by PPO. The generalist ($K=256$) is more efficient early on but the specialists are more effective in the end.}
\label{fig:obs1}
\end{figure}

A motivating observation for our framework is that training with more environment variations tends to be faster at the beginning but plateaus at a sub-optimal performance. We show the evidence here, by varying the number of environment variations to train an RL agent. We pay attention to both the learning speed (efficiency) and the performance at plateau (effectiveness).

On BigFish and StarPilot in Procgen, we use $256$ levels generated with seeds from $2000$ to $2256$. 
We choose $K=\{4,16,64,256\}$ for the number of levels per specialist. Notice that $K=256$ is equivalent to training a single generalist over all levels. For each $K$, we evenly and randomly distribute the $256$ training levels to $N_s=256/K$ specialists so that each specialist only learn on its distinctive set of levels. Each of the $N_s$ specialist has a budget of $200$ million/$N_s$ samples to train the PPO, so that the total sample budget is fixed for fair comparison of sample complexity. We use the default network architecture and hyper-parameters for PPO in the Procgen paper, except that we reduce the number of parallel threads from $64$ to $16$ for better sample efficiency for specialists of $K=\{4, 16\}$. We plot the training curve as the average of the $N_s=256/K$ specialists and scale it horizontally so that it reflects the total number of samples in a way same as the generalist's training curve. 

The results in Fig. \ref{fig:obs1} clearly suggest our findings. In short, when trained from scratch, training a generalist enjoys a more efficient early learning process; in contrast, later on, training specialists with smaller variations is more effective in achieving better performance without plateauing early.

\subsection{Influence of Specialist Training Timing on Efficiency and Effectiveness}
\vspace{-0.7em}
\begin{figure}[h]
\subfloat{\includegraphics[width=.45\linewidth]{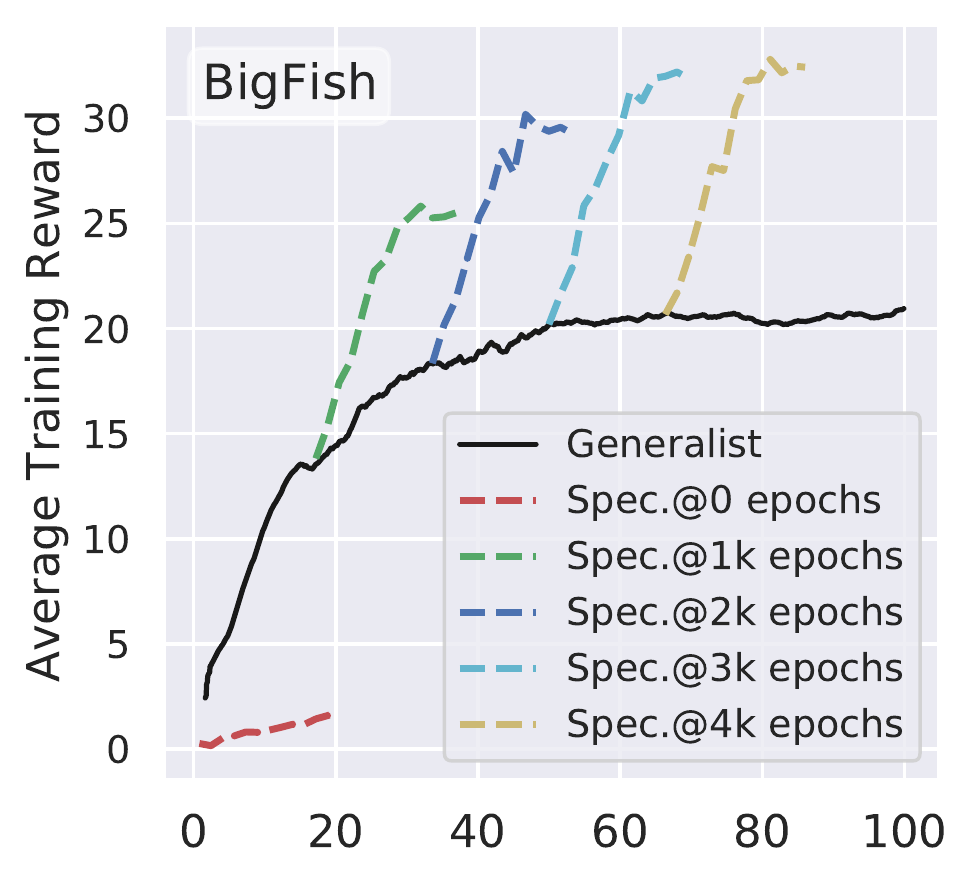}}
\subfloat{\includegraphics[width=.46\linewidth]{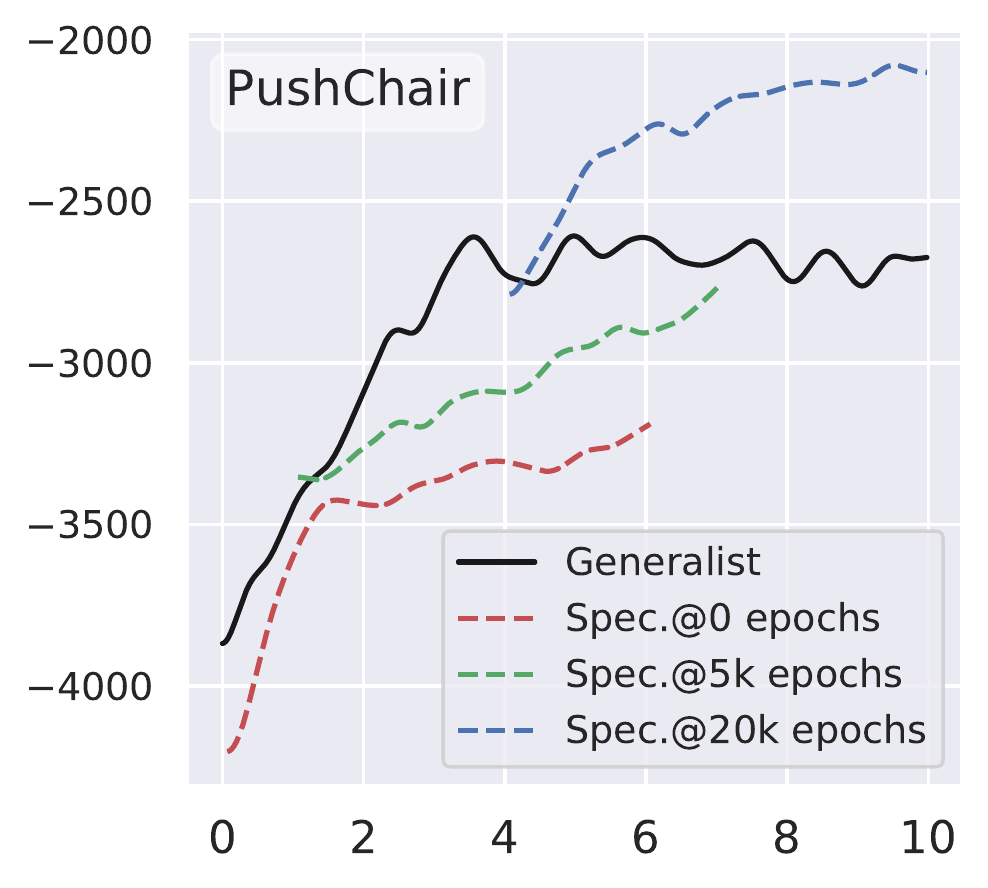}}\\
\vspace{-1.7em}
\caption{Training curves of specialists launched from the generalist trained with different numbers of epochs. Starting specialist training after generalist's performance plateaus is more sample efficient.}
\label{fig:specialist_timing}
\end{figure}

In Sec.~\ref{sec:when_how_train_specialists}, we mentioned that the timing to start specialist training plays a crucial role in the efficiency and effectiveness of our framework. In this section, we show further evidence. We launch specialist training at different stages of generalist's training curve and present results in Fig.~\ref{fig:specialist_timing}. We also conduct an experiment where specialists are trained from scratch instead of initiated from a generalist checkpoint (this corresponds to ``Spec @ 0 epochs'' in the figure). We use 64 specialist agents for BigFish training and 8 specialist agents for PushChair training.

We observe that, when the generalist is still in the early stages of fast learning and has not reached a performance plateau, launching specialist training results in worse sample complexity. In particular, training specialists from scratch as in previous works \citep{teh2017distral, ghosh2017divide} leads to inefficient and ineffective learning. On the other hand, after the generalist reaches a performance plateau, specialist training has very close efficiency and efficacy regardless of when the training is launched, as shown in the nearly parallel specialist curves in BigFish after the generalist has been trained for 2k epochs. Therefore, a good strategy for specialist training is to launch it as soon as the generalist reaches a performance plateau, which we adopt in Algorithm \ref{alg:gsl}.

\subsection{Hyperparameter Tuning of Plateau Criteria $\mathcal{H}$}

In Appendix, we show that our performance plateau criteria $\mathcal{H}$ (see Sec. \ref{sec:when_how_train_specialists}) can identify when the generalist reaches a performance plateau. Fig. \ref{fig:H} provides a qualitative evaluation on Procgen, ManiSkill and Meta-World. The smoothing kernel size is $200$ epochs for Procgen \& ManiSkill and $10$ for Meta-World; the window size is $W=600$ for Procgen, $W=2000$ for ManiSkill and $W=10$ for Meta-World; $\epsilon=0.03$ for Procgen, $\epsilon=0.01$ for Meta-World and $\epsilon=0.05k$ for ManiSkill. Areas around the black vertical lines are suitable for starting specialist training. 

Due to the different training dynamics of various RL algorithms on the benchmarks, the plateau criteria $\mathcal{H}$ does require some hyper-parameter tuning.
However, we find it relatively robust and easy to tune.
It turns out that starting specialist learning at a time $5\%$ of the total training budget earlier or later than the one suggested by $\mathcal{H}$ achieves very similar final results.
Nevertheless, if $\mathcal{H}$ is tuned too aggressively (too early), we lose some efficiency from the initial generalist learning; if it is tuned too conservatively (too late), the generalist plateaus too long (i.e., less efficiency) and could become too certain about its decisions, rendering it harder to be fine-tuned with specialists' knowledge.

\subsection{Diagnosis into Generalist Performance at Plateau}
\begin{figure}[t!]
\centering
\includegraphics[width=0.85\linewidth]{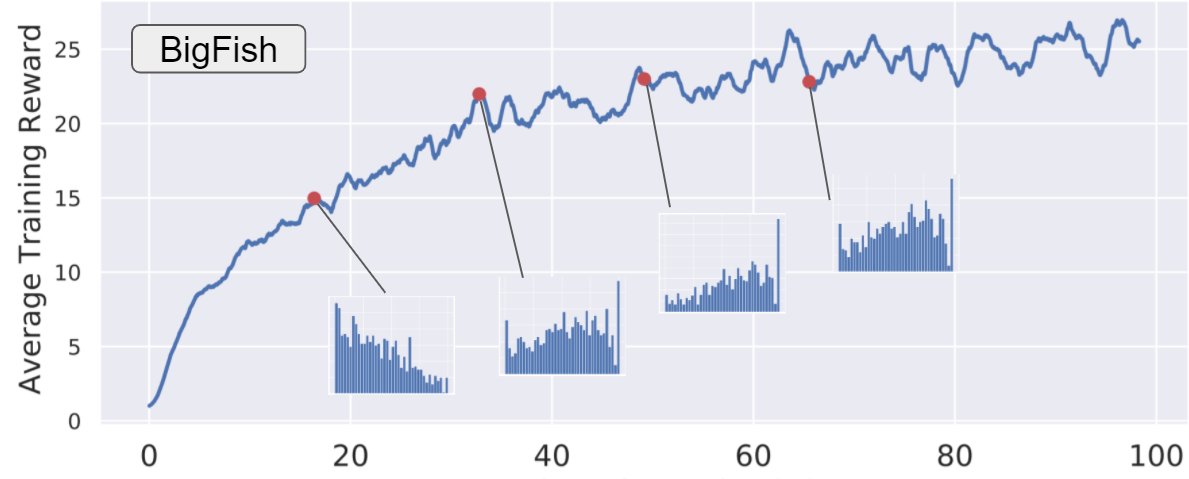}
\caption{Training returns of PPO on BigFish with 256 levels and performance histograms over the withheld levels at four generalist checkpoints during training. Policy learning on BigFish achieves good progress in a large portion of levels but slow to no improvements on the remaining. 
}
\label{fig:level_performance_distribution}
\end{figure}

\begin{table}[b]
\centering
\begin{tabular}{c|cc} 
\hline
& BigFish (train) & BigFish (test) \\
\hline
PPO (generalist only) & 24.6$\pm$0.7 & 24.3$\pm$1.1 \\
DnC RL & - & - \\
GSL with BC & 25.3$\pm$0.4 & 22.1$\pm$0.9 \\ 
GSL with DAPG & \textbf{31.1$\pm$0.8} & \textbf{30.0$\pm$0.5} \\
\hline
\end{tabular}
\caption{Comparison to baselines and other design choices. DnC RL requires synchronized training across all specialists with a quadratic computational complexity w.r.t. the number of specialists, and we find it hard to scale up to our setup of 75 specialists as in GSL. Using GSL with BC harms the performance. Our design works the best over all choices.}
\label{tb:LfD}
\end{table}

In Sec.~\ref{sec:when_how_train_specialists}, we discussed the strategy to assign environment variations to specialists. 
Specifically, we launch specialist training only on the $N_s$ variants with the lowest training performance evaluated using the generalist.
Our strategy is based on the observation that the generalist can solve some environment variations reasonably well, but performs poorly on others. 
We visualize performance of the generalist on BigFish from Procgen in Fig.~\ref{fig:level_performance_distribution}, where the PPO agent makes quick progress on most of the levels, with poor performance only on a portion of variations at the time of plateau.
We find this quite common for environments in Procgen, except for Plunder where PPO struggles in a majority of levels.

\subsection{Influence of Specialist-to-Generalist Algorithm on Final Performance}
In Sec.~\ref{sec:specialist2generalist}, we mentioned the importance of selecting the algorithm for consolidating specialists' experience for generalist learning. We conduct experiments on BigFish of Procgen, where we set the number of specialists to be 75. We compare GSL+BC, GSL+DAPG, and DnC~\citep{ghosh2017divide}, a classic divide-and-conquer RL work which uses BC. As a baseline, we also show the performance of jointly training a PPO on all variations. However, for DnC RL, we find that unlike any other setups, it has quadratic computational complexity w.r.t. the number of specialists, rendering it computationally intractable, so we skip this method. As shown in Table.~\ref{tb:LfD}, using BC in our GSL framework does not help to outperform a single PPO generalist. One possible reason is that GSL does not restrict specialists to be close to the generalist as in DnC, so specialists' demonstrations can be inconsistent, posing learning difficulties for BC. On the other hand, our GSL with DAPG can outperform a single PPO by a large margin. Additionally, we sample 1,000 novel levels from BigFish to test the generalization performance of our models. Results show that GSL+DAPG stays the best for generalization.




%% file: sections/conclusion.tex
\section{Conclusion}
Generalization in RL usually requires training on diverse environments. In this work, we develop a simple yet effective framework to solve RL problems that involve a large number of environment variations. The framework is a meta-algorithm that turns a pair of RL algorithm and learning-from-demonstrations algorithm into a more powerful RL algorithm. By analysis into prototypical cases, we identify that the catastrophic ignorance and forgetting of neural networks pose significant challenges to RL training in environments with many variations, and may cause the agent to reach performance plateau at sub-optimal level. We show that introducing specialists to train in subsets of environments can effectively escape from this performance plateau and reach a high reward. Design choices that are crucial yet unknown in the literature must be taken care of for the success of our framework. Empirically, our framework achieves high efficiency and effectiveness by improving modern RL algorithms on several popular and challenging benchmarks.

%% file: sections/appendix.tex
\section{Hyperparameters}
\label{app:hyper-all}

\subsection{Illustrative example}
\label{app:ill}
In our illustrative example, we train the generalist and the specialists using PPO, and use DAPG + PPO to resume generalist training with demos from specialists. We use the following hyperparameters:

\begin{table}[h]
\centering
\small
\begin{tabular}{r|cc}
\toprule
Hyperparameters & Value\\
\midrule
Optimizer & Adam \\
Learning rate & $3 \times 10^{-4}$ \\
Discount ($\gamma$) & 0.95 \\
$\lambda$ in GAE & 0.97 \\
PPO clip range & 0.2 \\
Coefficient of the entropy loss term of PPO $c_{ent}$ & 0.01 \\
Coefficient of the entropy loss term of PPO $c_{ent}$ (during DAPG) & 0.01 \\
Number of hidden layers (all networks) & 2\\
Number of hidden units per layer & 256\\
Number of threads for collecting samples & 5\\
Number of samples per PPO step & $10^4$\\
Number of samples per minibatch & 2000\\
Nonlinearity & ReLU \\
Total Simulation Steps & $5\times 10^6$\\
Number of environment variations & 5 \\
Environment horizon & 150 \\
\bottomrule
\end{tabular}
\vspace{1em}
\caption{Hyperparameters for DAPG and PPO in our illustrative example}
\label{tab:ill parameters}
\end{table}

\begin{table}[h]
\centering
\small
\begin{tabular}{r|cc}
\toprule
Hyperparameters & Value\\
\midrule
Number of specialists $N_s$ & $5$ \\
Number of lowest performing environment variations assigned to specialists $N_{lenv}$ & $5$ \\
Number of environment variations per specialist $K$ & $1$ \\
Number of demo samples from specialists (epochs $\times$ time steps per epoch $\times$ $N_s$) $N_D^s$ & $10 \times 150 \times 5$ \\
Number of demo samples from generalist $N_D^g$ & N/A \\
Sliding window size (epochs) for the plateau criteria $W$ & $50$ \\
Number of samples for training each specialist $N_{sample}$ & $ 500K$ \\ 
Number of samples for DAPG & $1M$ \\
Margin in the plateau criteria $\epsilon$ & 0.01 \\
\bottomrule
\end{tabular}
\vspace{1em}
\caption{Hyperparameters of GSL in our illustrative example.}
\label{tab:ill GSL}
\end{table}

\subsection{Procgen}
\label{app:procgen}


In Procgen, we train the generalist and the specialists using PPO/PPG (which is an extension of PPO), and use DAPG + PPO/PPG to resume generalist training with demos from specialists. We use the following hyperparameters:

\begin{table}[H]
\centering
\small
\begin{tabular}{r|cc}
\toprule
Hyperparameters & Value\\
\midrule
Optimizer & Adam \\
Learning rate & $5 \times 10^{-4}$ \\
Discount ($\gamma$) & 0.999 \\
$\lambda$ in GAE & 0.95 \\
PPO clip range & 0.2 \\
Coefficient of the entropy loss term of PPO $c_{ent}$ & 0.01 \\
Coefficient of the entropy loss term of PPO $c_{ent}$ (during DAPG) & 0.05 \\
Number of threads for collecting samples & 64\\
Number of samples per PPO epoch & $256 \times 64$\\
Number of samples per minibatch & $1024$\\
Nonlinearity & ReLU \\
Total Simulation Steps & $10^8$\\
\bottomrule
\end{tabular}
\vspace{1em}
\caption{The hyperparameters of PPO and DAPG for Procgen experiments.}
\label{tab:Procgen parameters}
\end{table}

\begin{table}[H]
\centering
\small
\begin{tabular}{r|cc}
\toprule
Hyperparameters & Value\\
\midrule
Number of policy update epochs in each policy phase $\texttt{n\_pi}$ & 32 \\
Number of auxiliary epochs in each auxiliary phase $\texttt{n\_aux\_epochs}$ & 6 \\
Coefficient of the entropy loss term of PPO $c_{ent}$ & 0.0 \\
Coefficient of the entropy loss term of PPO $c_{ent}$ (during DAPG) & 0.01 \\
\bottomrule
\end{tabular}
\vspace{1em}
\caption{Additional hyperparameters of PPG for Procgen experiments.}
\label{tab:Procgen parameters}
\end{table}

\begin{table}[H]
\centering
\small
\begin{tabular}{r|cc}
\toprule
Hyperparameters & Value\\
\midrule
Number of specialists $N_s$ & $75$ \\
Number of lowest performing environment variations assigned to specialists $N_{lenv}$ & $300$ \\
Number of environment variations per specialist $K$ & $4$ \\
Number of demo samples (epochs $\times$ sampled time steps per epoch $\times N_s$) $N_D^s$ & $256\times 32 \times 75$ \\
Number of demo samples (epochs $\times$ sampled time steps per epoch $\times$ number of env. variations) $N_D^g$ & $256\times 8 \times 724$ \\
Sliding window size (epochs) for the plateau criteria $W$ & $600$ \\
Threshold for filtering demos (in terms of the normalized score) $\tau$ & $0.15$ \\
Number of samples for training each specialist $N_{sample}$ (PPO) & $16\times 256\times 64$ \\
Number of samples for training each specialist $N_{sample}$ (PPG) & $20\times 256\times 64$ \\
Number of samples for DAPG (for PPO) & $50 \times 256 \times 64$ \\
Number of samples for DAPG (for PPG) & $800 \times 256 \times 64$ \\
Margin in the plateau criteria $\epsilon$ & 0.03 \\
\bottomrule
\end{tabular}
\vspace{1em}
\caption{The hyperparameters of GSL for experiments on Procgen.}
\label{tab:Procgen-GSL}
\end{table}

\paragraph{Other implementation details} For all environments, we use seeds (levels) from $1000$ to $2023$ for training and from $100000$ to $100999$ for testing.
When training the specialists, we change the number of parallel environments in PPO from $64$ to $16$ for a slightly better sample efficiency, and we change both $\texttt{n\_pi}$ and $\texttt{n\_aux\_epochs}$ to 4 from PPG for a similar reason.
We find that increasing the coefficient $c_{ent}$ for the entropy regularization loss during DAPG can help improve both the optmization and generalization performance for Procgen.
For the Plunder environment (with PPO as the baseline), we have observed poor generalist training performance across a majority of levels. We therefore set $N_s=1024$, i.e., we train specialists on all levels. We also change the number of samples used in DAPG to $200\times 256\times 64$ since there are more demos collected from the specialists than that in other tasks. 
Moreover, in Plunder we set $c_{ent}$ to $0.1$ during DAPG.

\subsection{Meta-World}

For both MT-10 and MT-50 from Meta-World, we train the generalist and the specialists using PPO (with a policy network of two hidden layers, each of hidden size 32), and use DAPG + PPO to resume generalist training with demos from specialists. We use the following hyperparameters:

\begin{table}[H]
\centering
\small
\begin{tabular}{r|cc}
\toprule
Hyperparameters & Value\\
\midrule
Optimizer & Adam \\
Learning rate & $2.5 \times 10^{-4}$ \\
Discount ($\gamma$) & 0.99 \\
$\lambda$ in GAE & 0.95 \\
Minimum std. of the Gaussian policy $\texttt{min\_std}$ & 0.5 \\
Maximum std. of the Gaussian policy $\texttt{max\_std}$ & 1.5 \\
PPO clip range & 0.2 \\
Coefficient of the entropy loss term of PPO $c_{ent}$ (MT-10) & 0.005 \\
Coefficient of the entropy loss term of PPO $c_{ent}$ (MT-50) & 0.05 \\
Number of threads for collecting samples (MT-10) & 10 \\
Number of threads for collecting samples (MT-50) & 50 \\
Number of samples per PPO epoch & $10^5$\\
Number of samples per minibatch & $32$\\
Nonlinearity & ReLU \\
Total Simulation Steps (MT-10) & $2\times 10^7$\\
Total Simulation Steps (MT-50) & $4\times 10^7$\\
\bottomrule
\end{tabular}
\vspace{1em}
\caption{The hyperparameters of PPO and DAPG for Meta-World experiments.}
\label{tab:Procgen parameters}
\end{table}

\begin{table}[H]
\centering
\small
\begin{tabular}{r|cc}
\toprule
Hyperparameters & Value\\
\midrule
Number of lowest performing environment variations assigned to specialists $N_{lenv}$ & Varied \\
Number of specialists $N_s$ & $N_{lenv}$ \\
Number of environment variations per specialist $K$ & $1$ \\
Number of demo samples (sampled time steps per env. variations $\times N_s$) $N_D^s$ & $10^5 \times N_s$ \\
Number of demo samples (sampled time steps per env. variations $\times$ number of remaining env. variations $N_r$) $N_D^g$ & $10^5 \times$ $N_r$ \\
Sliding window size (epochs) for the plateau criteria $W$ & $10$ \\
Threshold for filtering demos (in terms of success rate) $\tau$ & $1.0$ (successful) \\
Number of samples for training each specialist $N_{sample}$ & $3000\times 700$ \\
Number of samples for DAPG (MT-10) & $2\times 10^6$ \\
Number of samples for DAPG (MT-50) & $6\times 10^6$ \\
Margin in the plateau criteria $\epsilon$ & 0.01 \\
\bottomrule
\end{tabular}
\vspace{1em}
\caption{The hyperparameters of GSL for experiments on Meta-World.}
\label{tab:Procgen-GSL}
\end{table}

\paragraph{Other implementation details} We find that in MT-10/50, there always exist some tasks (i.e., environment variations) that the generalist agent performs extremely poorly after the initial learning phase. We therefore use a threshold of success rate of $0.5$ for MT-10 and $0.2$ for MT-50 to select the $N_{lenv}$ (which varies across different runs) lowest performing env. variations and correspondingly launch $N_{lenv}$ specialists. 
During specialist training, we reduce the number of samples per PPO epoch from $10^5$ to $3\times 10^3$ to improve the sample efficiency (since each specialist now only learns to solve one environment variation).

\subsection{ManiSkill}
\label{app:maniskill}
For ManiSkill experimenets, we adopt the same PointNet + Transformer over object segmentation architecture as in the original paper. We proportionally downsample point cloud observations to 1200 points following the same strategy in the original paper. We train the generalist and the specialists using SAC, and use GAIL + SAC to resume generalist training with demos from specialists. Notice that each specialist only focuses on one chair model in the PushChair task. We use the hyperparameters listed in Tab. \ref{tab:SAC parameters}, inspired by the implementations of~\citet{shen2022learning}. We also show hyperparameters for GSL in Tab. \ref{tab:mainiskill-gsl}.

\begin{table}[t]
\centering
\small
\begin{tabular}{r|cc}
\toprule
Hyperparameters & Value\\
\midrule
Optimizer & Adam \\
Learning rate & $3 \times 10^{-4}$ \\
Discount ($\gamma$) & 0.95 \\
Replay buffer size ($\gamma$) & $2 \times 10^6$ \\
Number of threads for collecting samples & 4\\
Number of samples per minibatch & 200\\
Nonlinearity &ReLU \\
Target smoothing coefficient($\tau$) & 0.005 \\
Target update interval & 1 \\
$Q$, $\pi$ update frequency & 4 updates per 64 online samples \\
GAIL discriminator update frequency & 5 updates per 100 policy updates \\
Total Simulation Steps & $2\times 10^7$\\
\bottomrule
\end{tabular}
\vspace{1em}
\caption{The hyperparameters of SAC and GAIL+SAC for ManiSKill}
\label{tab:SAC parameters}
\end{table}

\begin{table}[t]
\centering
\small
\begin{tabular}{r|cc}
\toprule
Hyperparameters & Value\\
\midrule
Number of specialists $N_s$ & $8$ \\
Number of lowest performing environment variations assigned to specialists $N_{lenv}$ & $8$ \\
Number of environment variations per specialist $K$ & $1$ \\
Number of demo samples from specialists $N_D^s$ & $200\times 300\times 8$ \\
Number of demo samples from generalist $N_D^g$ & N/A \\
Sliding window size (epochs) for the plateau criteria $W$ & $2000$ \\
Threshold for filter the demos $\tau$ & $-3.5 \times 10^3$ \\
Number of samples for training each specialist $N_{sample}$ & $1.5\times 10^6$ \\ 
Number of samples used in GAIL+SAC & $2\times 10^6$ \\
Margin used in the plateau criteria $\epsilon$ & 50 \\
\bottomrule
\end{tabular}
\vspace{1em}
\caption{The hyperparameters of GSL for experiments on ManiSkill.}
\label{tab:mainiskill-gsl}
\end{table}

\section{Criteria function $\mathcal{H}$ for performance plateaus}

\begin{figure}[h]
\centering
\begin{subfigure}[b]{\linewidth}
\centering
\includegraphics[width=0.195\linewidth]{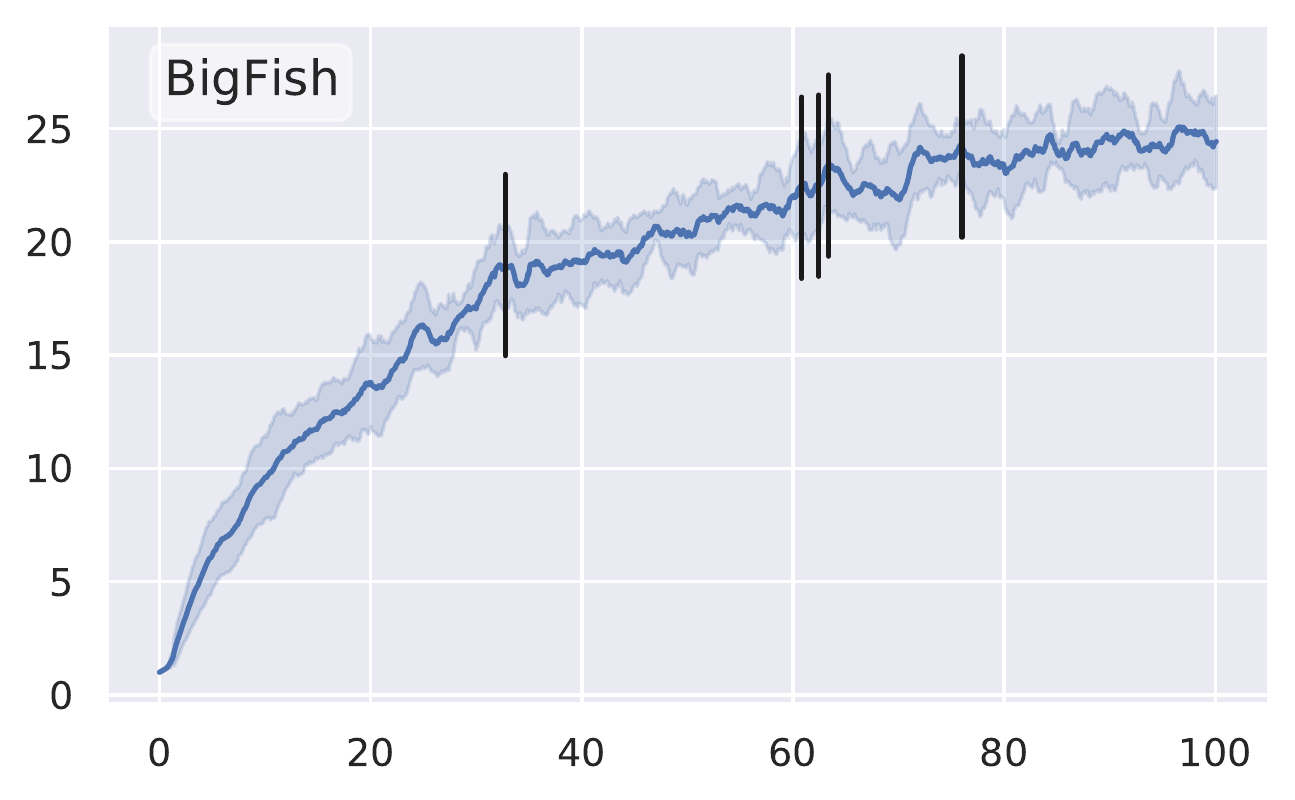}
\includegraphics[width=0.195\linewidth]{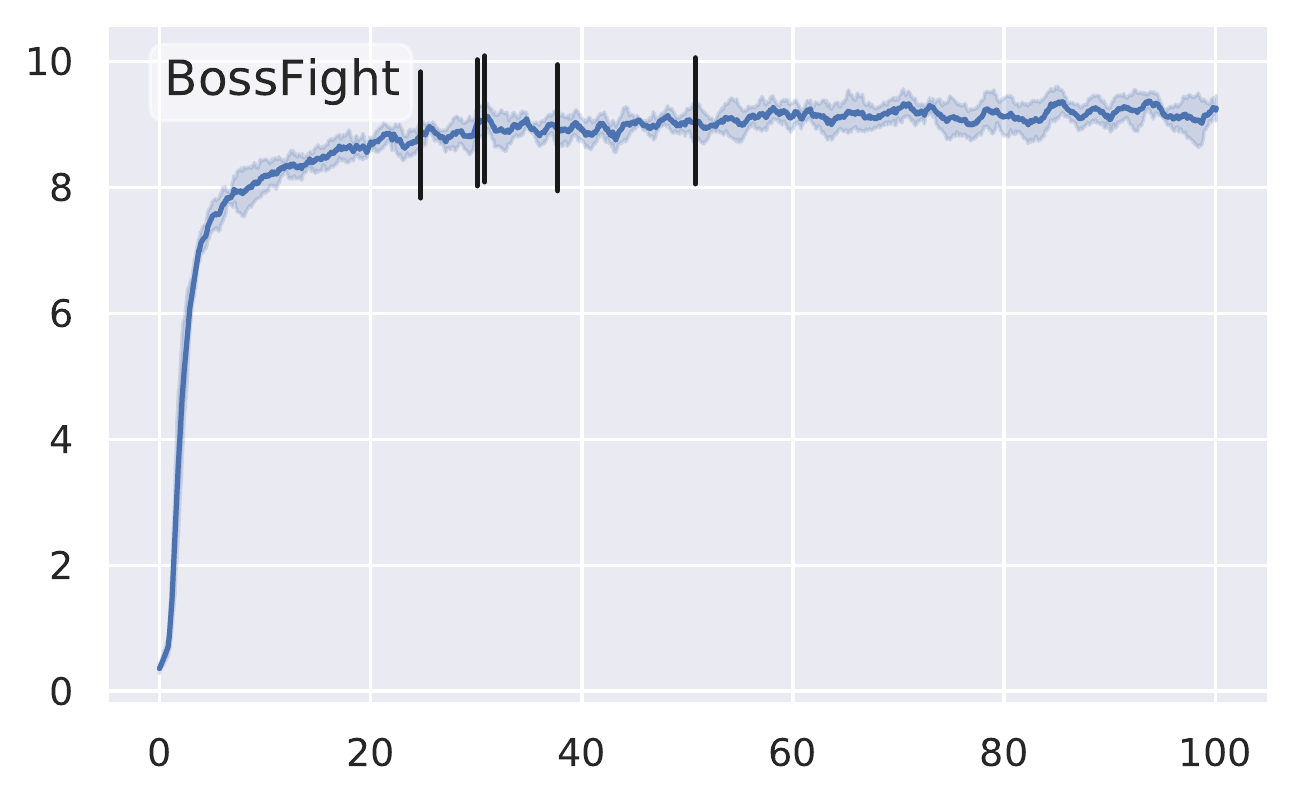}
\includegraphics[width=0.195\linewidth]{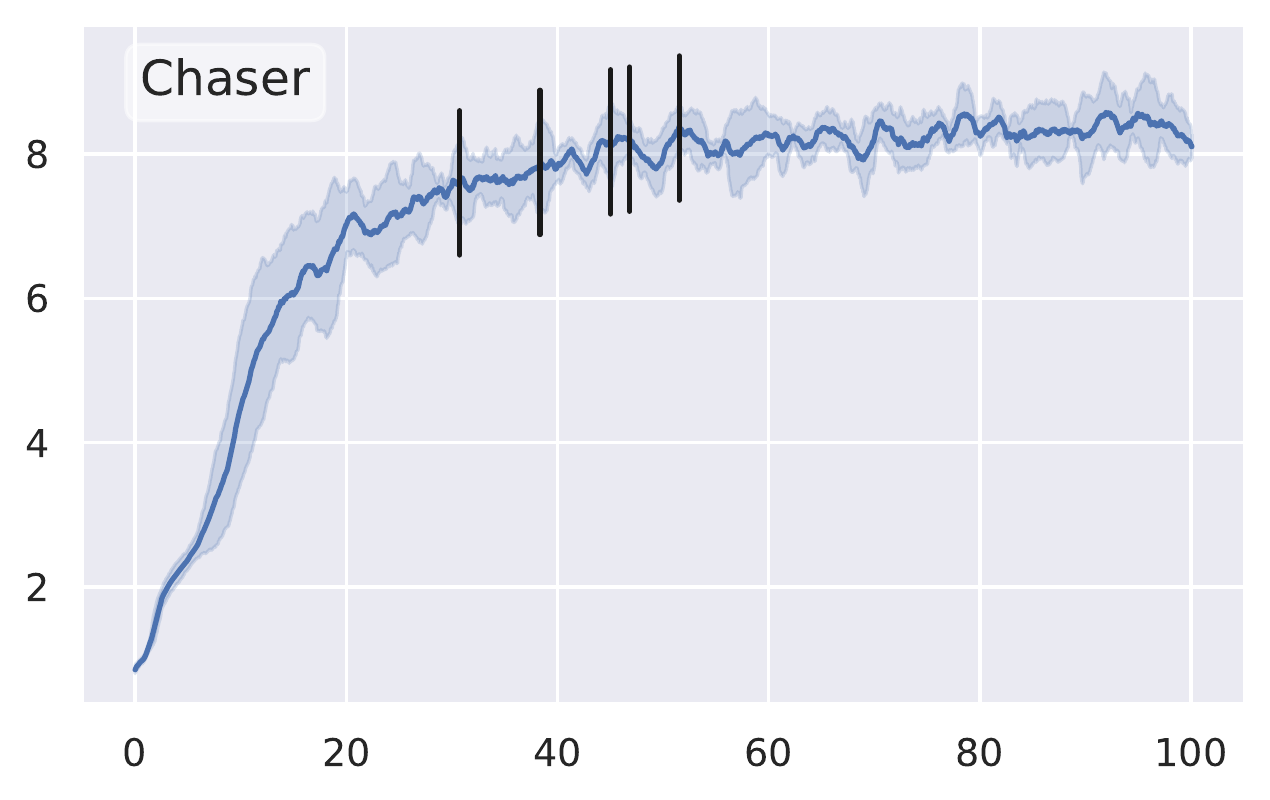}
\includegraphics[width=0.195\linewidth]{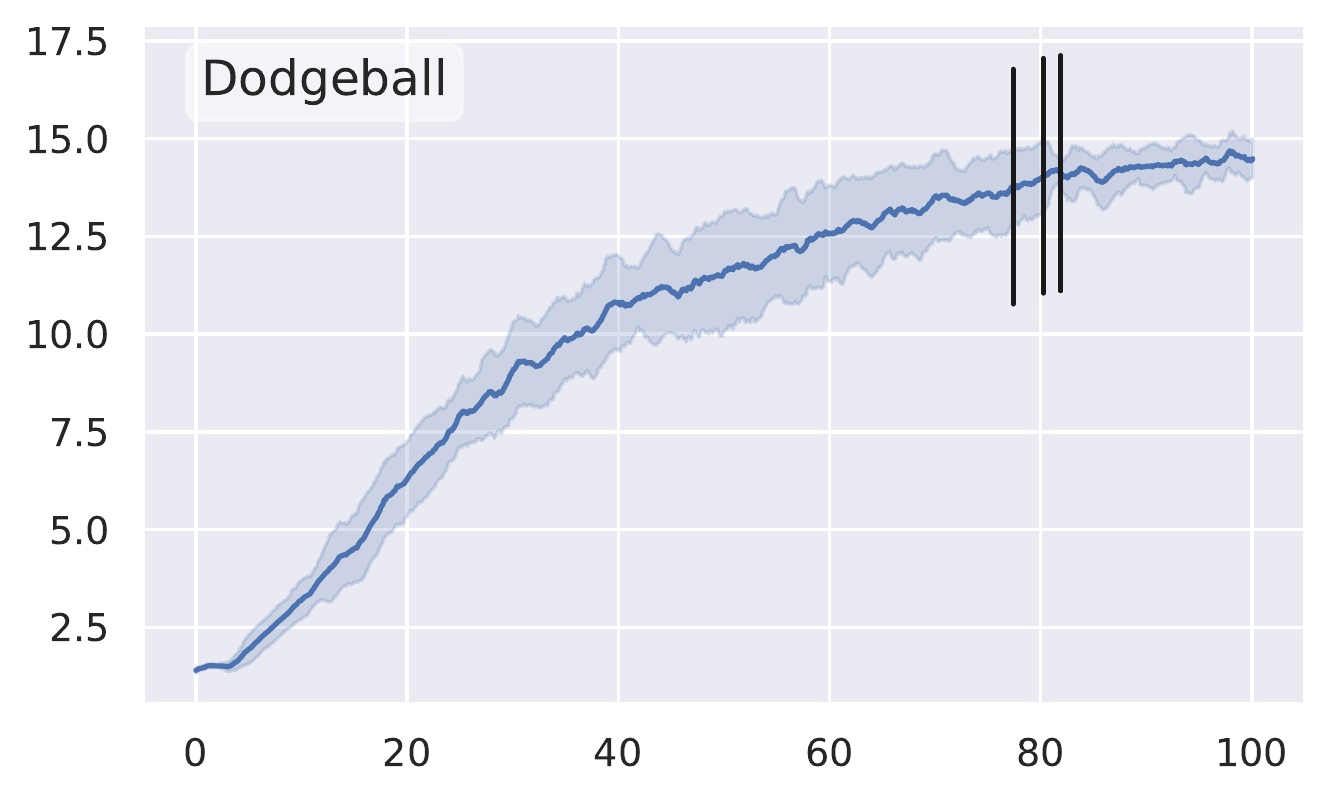}
\includegraphics[width=0.195\linewidth]{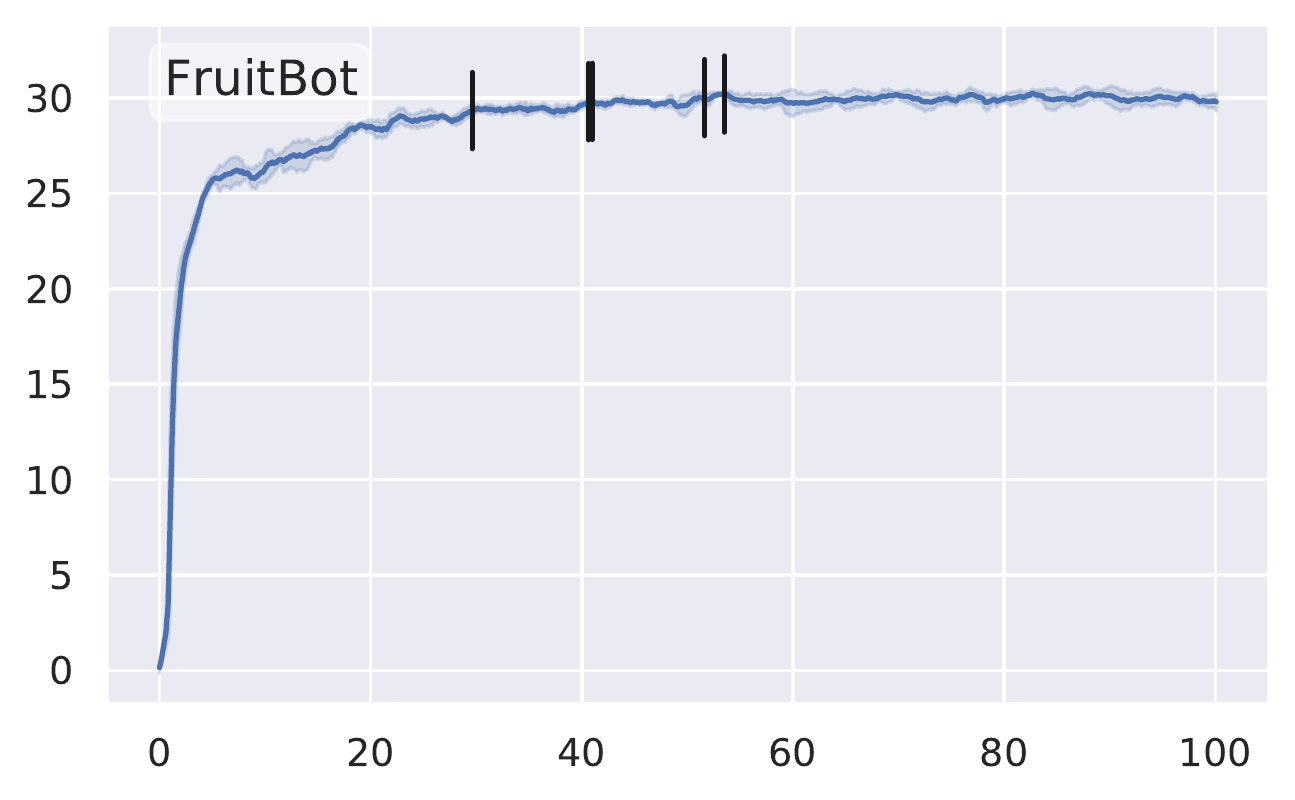}
\end{subfigure}
\begin{subfigure}[b]{\linewidth}
\centering
\includegraphics[width=0.195\linewidth]{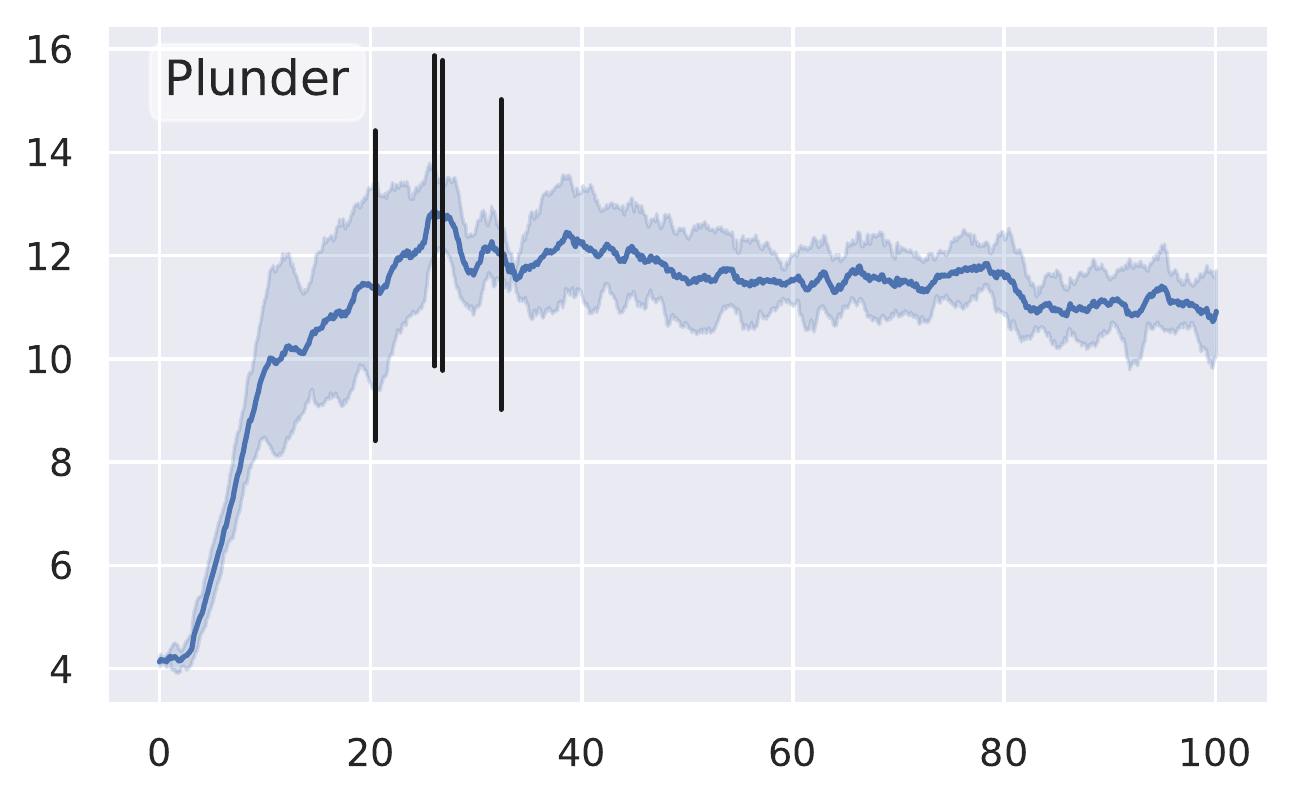}
\includegraphics[width=0.195\linewidth]{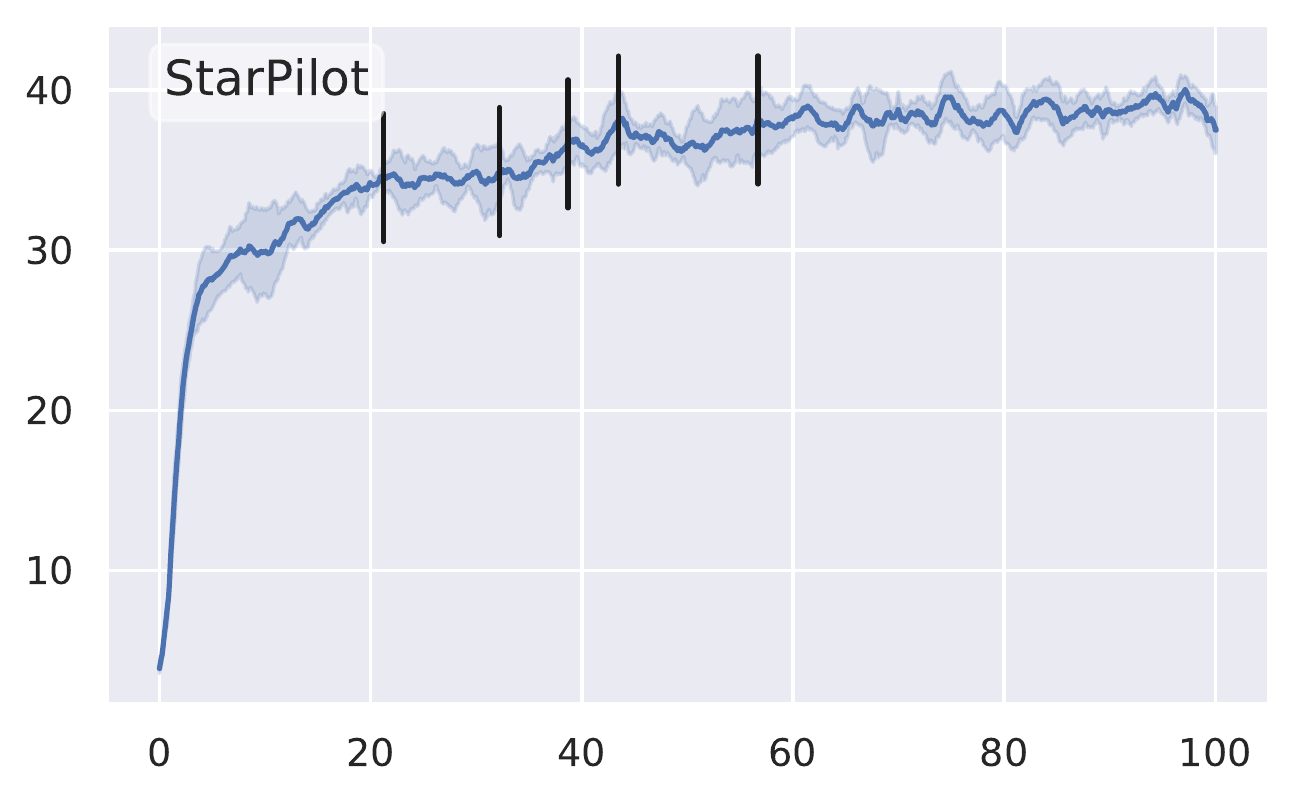}
\includegraphics[width=0.195\linewidth]{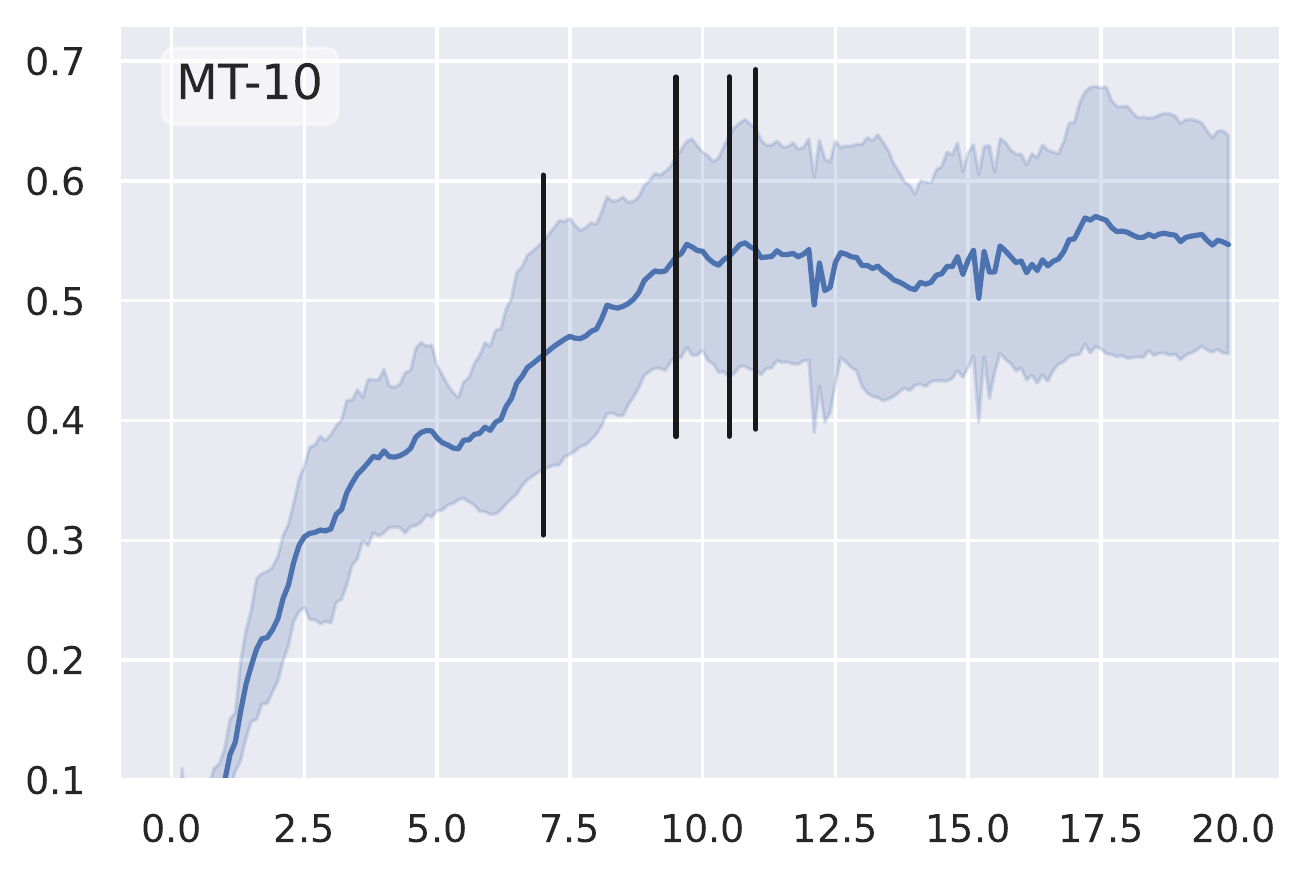}
\includegraphics[width=0.195\linewidth]{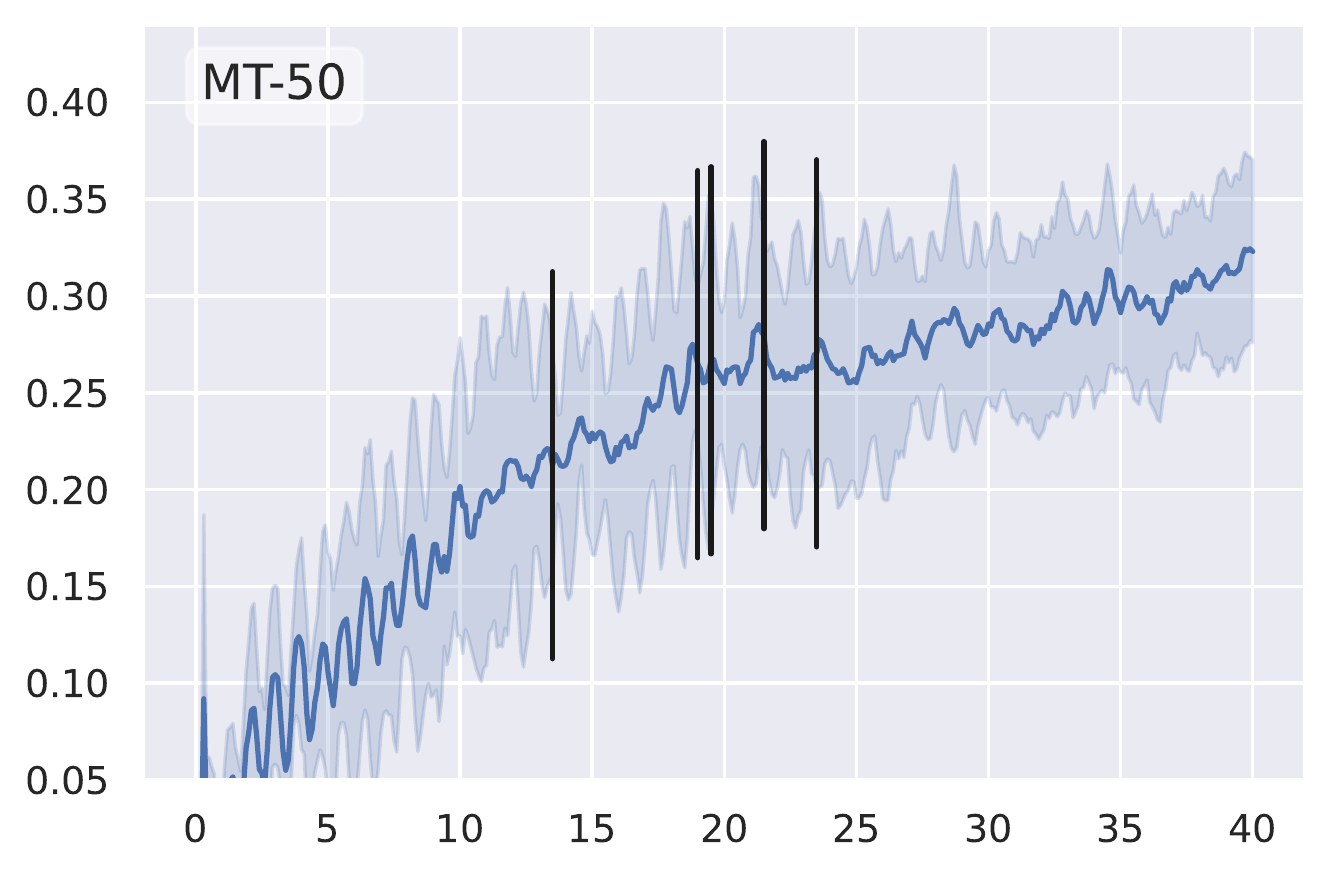}
\includegraphics[width=0.195\linewidth]{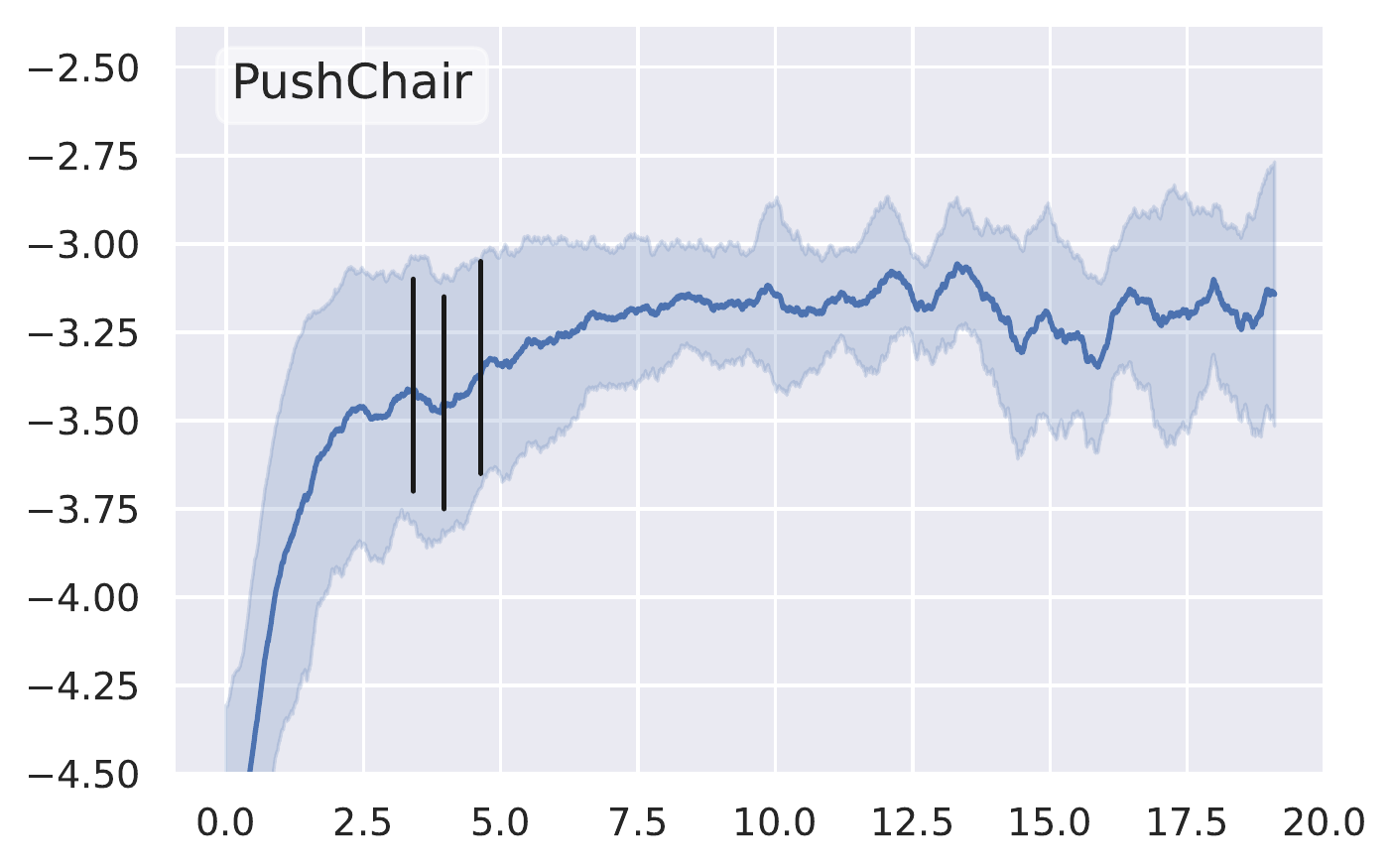}
\end{subfigure}
\caption{Qualitative evaluation of criteria $\mathcal{H}$. Vertical black lines indicate where $\mathcal{H}(t) = 1$ for the first time in each run of $5$ runs for Procgen and Meta-World and 3 runs for PushChair in ManiSkill. The y-axis is raw episode rewards for Procgen and ManiSkill (whose unit is 1000) and average success rate for Meta-World. The x-axis is million (of steps).}
\label{fig:H}
\end{figure}

Here we verify the effectiveness of our criteria $\mathcal{H}$ that indicates when the generalist's training performance plateaus.
Specifically, for each of the 5 runs in Procgen environments and one run in PushChair from ManiSkill, we mark the first timestamp $t$ where our proposed criteria $\mathcal{H}(t)$ is evaluated as 1. We qualititatively demonstrate the effectiveness of our criteria in Fig. \ref{fig:H}.

In addition, in practice, we do not consider the first 15\% to avoid the cases where the generalist gets stuck at the very beginning (i.e., a ``slow start'' generalist).
We also do not consider the last 15\% to leave a sufficient amount of samples for launching specialist training and using specialists' demonstrations to guide generalist training.

\section{Training curves of multiple runs}

For the main results on Procgen and Meta-World, we perform 5 runs for both the baseline and GSL in each environment; for ManiSkill, we perform 3 runs (due to its high computation complexity). 
In the main paper, we only plot one run of training curve for GSL (for each environment) to ensure a clear presentation.
Here, we aggregate the curves across multiple runs to illustrate the mean rewards and their standard deviations.
Notice that, due to the performance plateau criteria $\mathcal{H}$, each run has different starting and ending points for both the generalists and the specialists.
We therefore normalize the x-axis to align each training curve.
Specifically, we use percentage (ranging from 0 to 100) as the x-axis.
We display curves for the initial generalist training in the first 50\%, those for the specialists in the next 25\% and finally the fine-tuned generalist in the remaining 25\%.

Since the training curves plotted here are subject to smoothing, their values (mean episode rewards or success rates) at the end of the training process might not exactly equal the results reported in Tab. \ref{tab:main_procgen_train} \& \ref{tab:extra}, which are obtained by batch evaluations using the model checkpoints.

\begin{figure*}
\begin{subfigure}[b]{\linewidth}
    \begin{subfigure}[b]{\linewidth}
        \centering
        \includegraphics[width=0.24\linewidth]{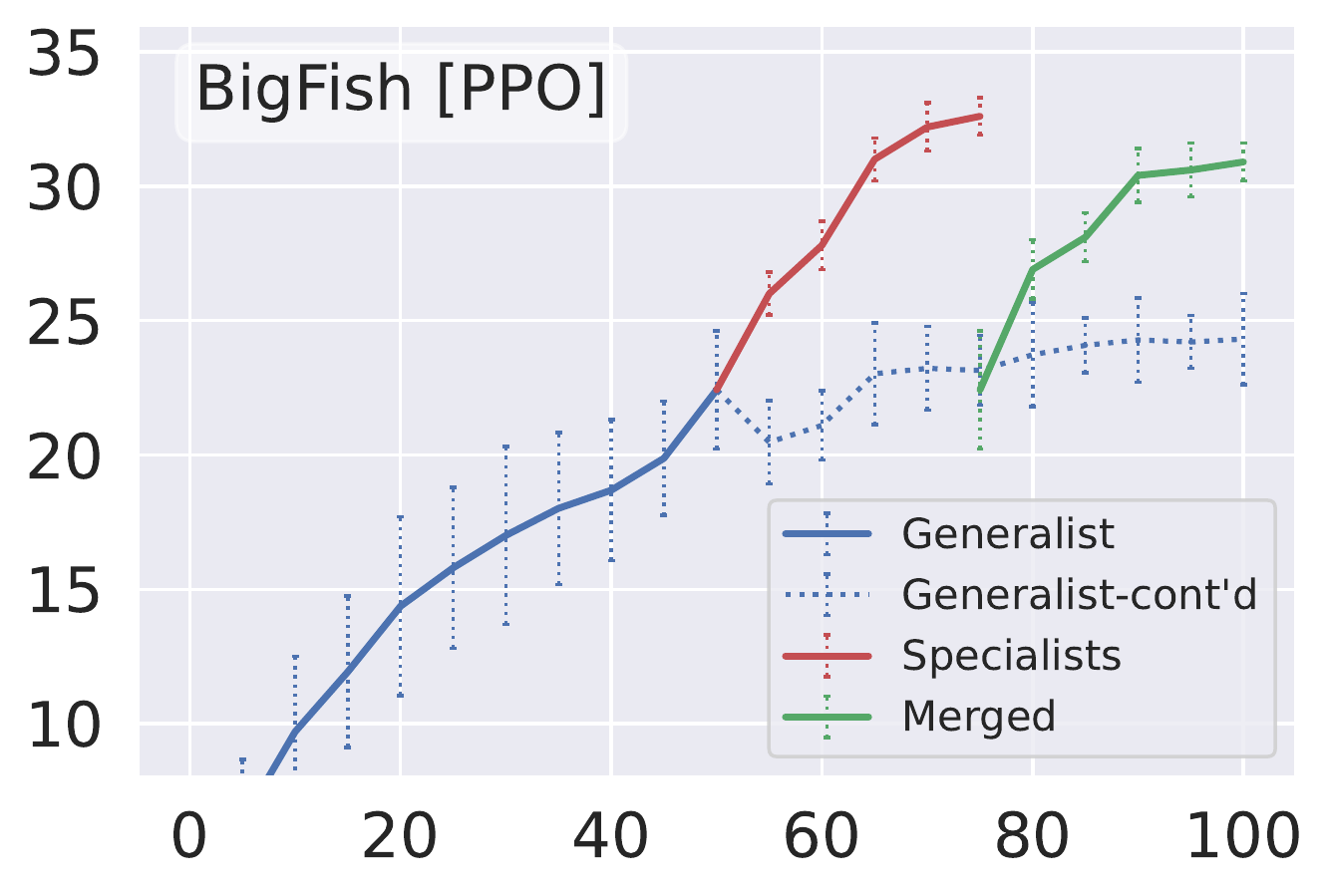}
        \includegraphics[width=0.24\linewidth]{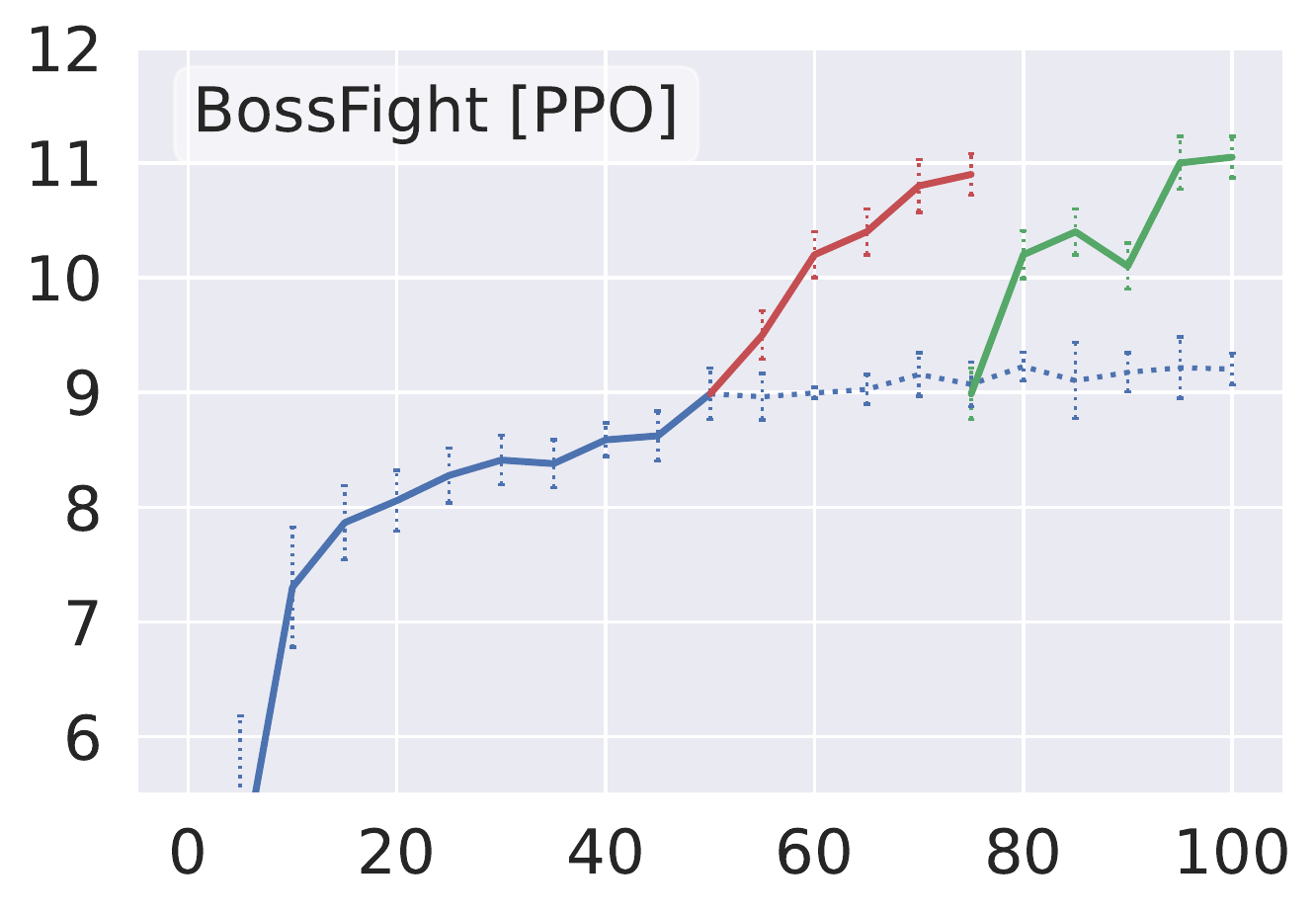}
        \includegraphics[width=0.24\linewidth]{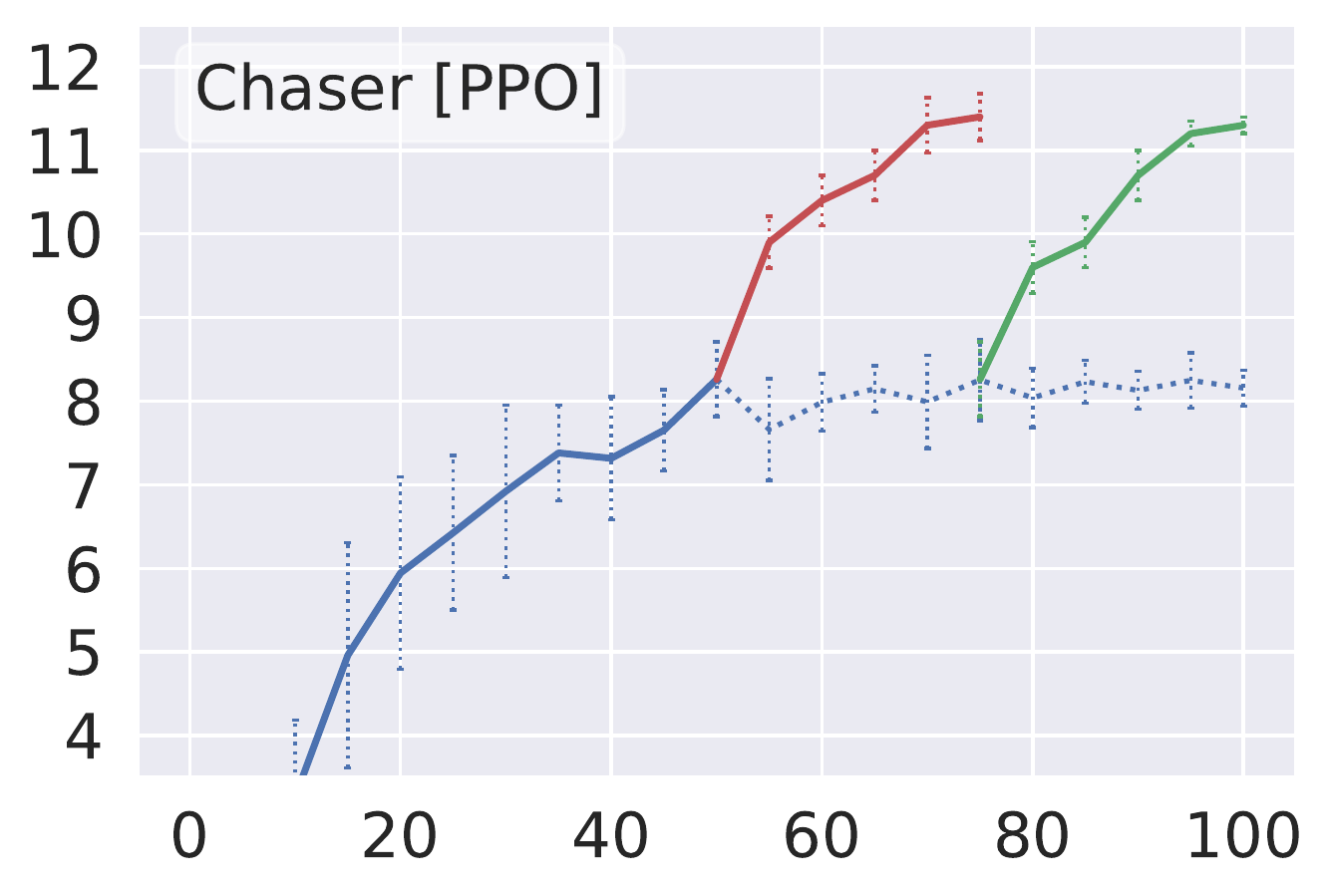}
        \includegraphics[width=.24\linewidth]{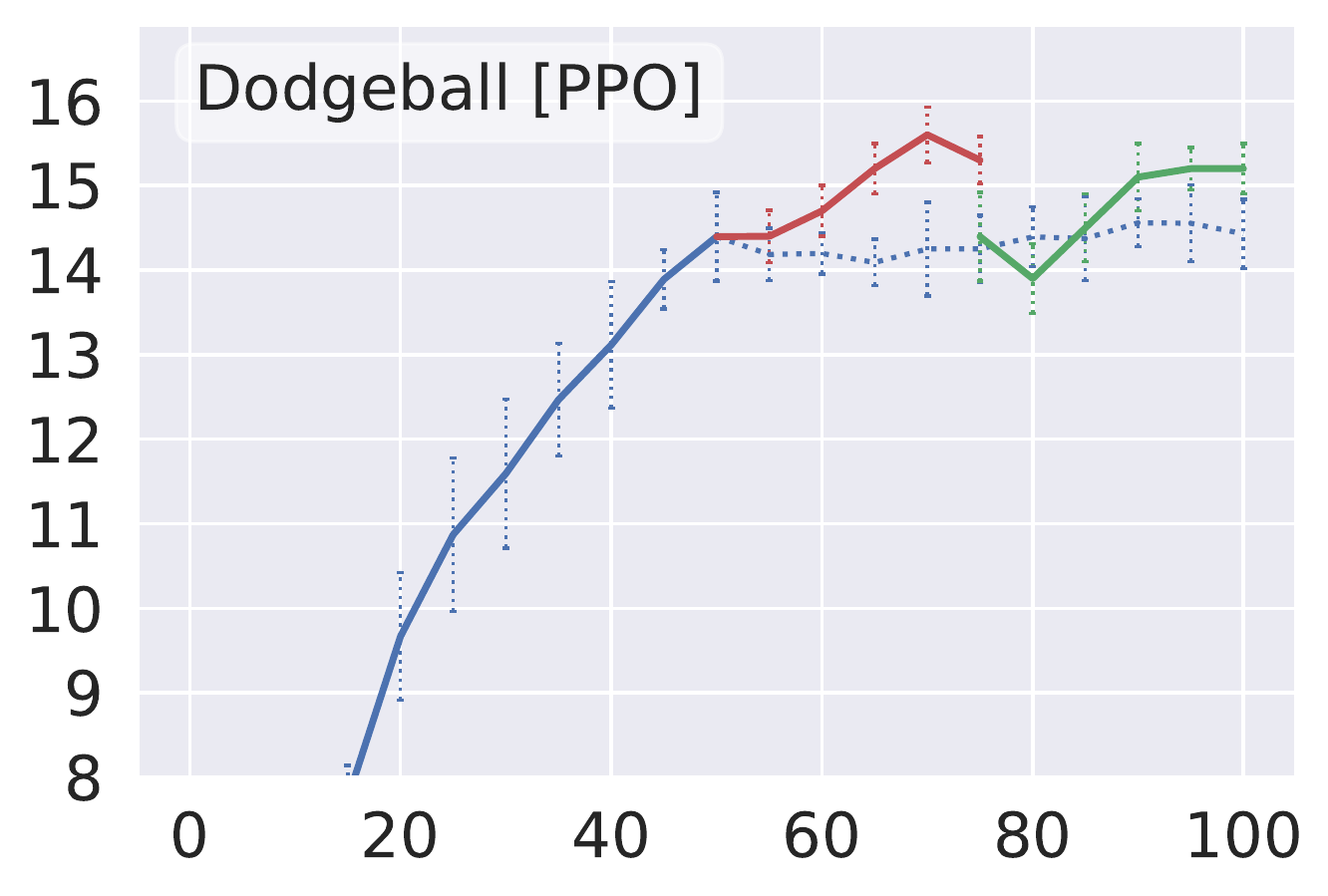}
    \end{subfigure}
    \begin{subfigure}[b]{\linewidth}
        \centering
        \includegraphics[width=0.24\linewidth]{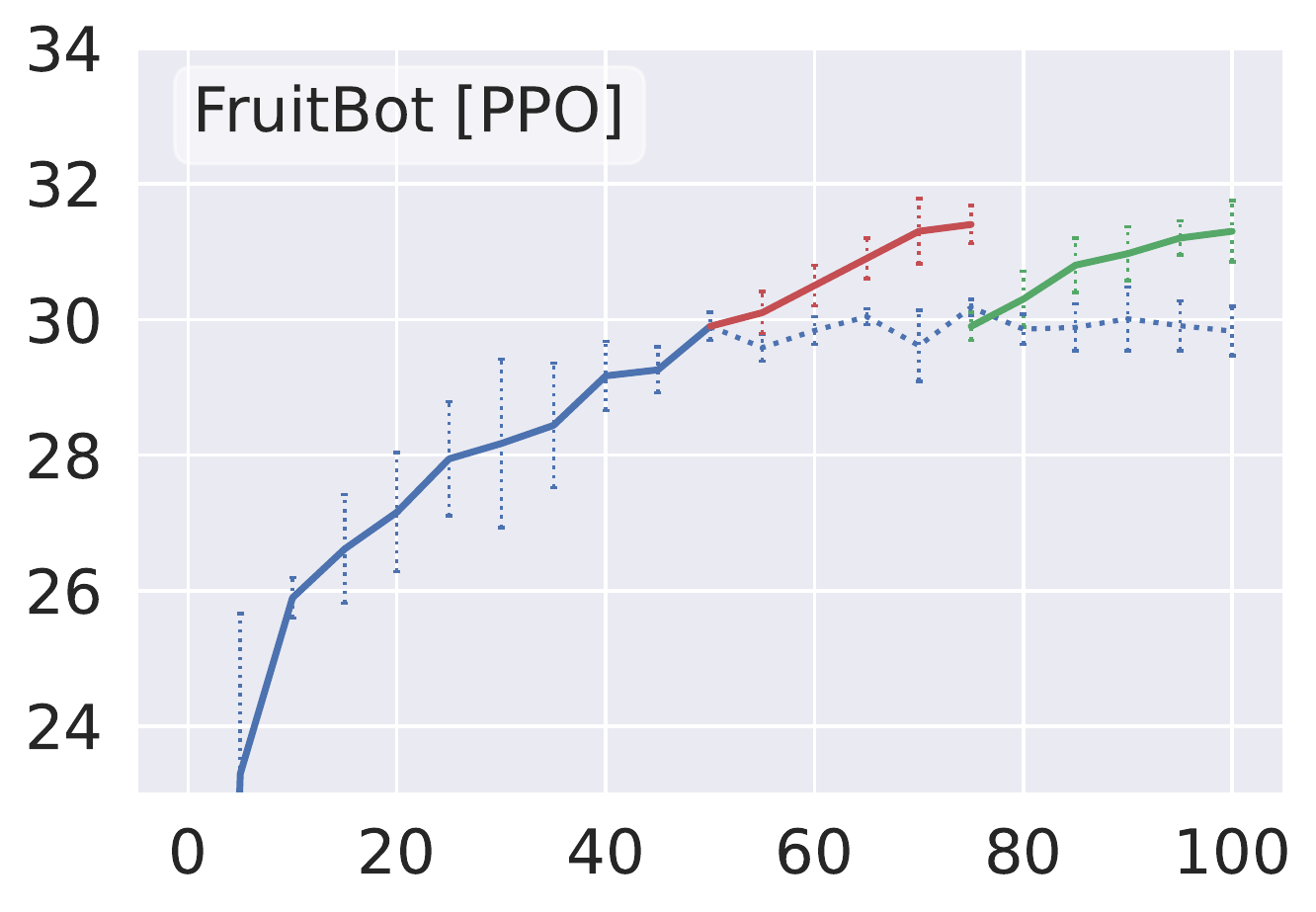}
        \includegraphics[width=0.24\linewidth]{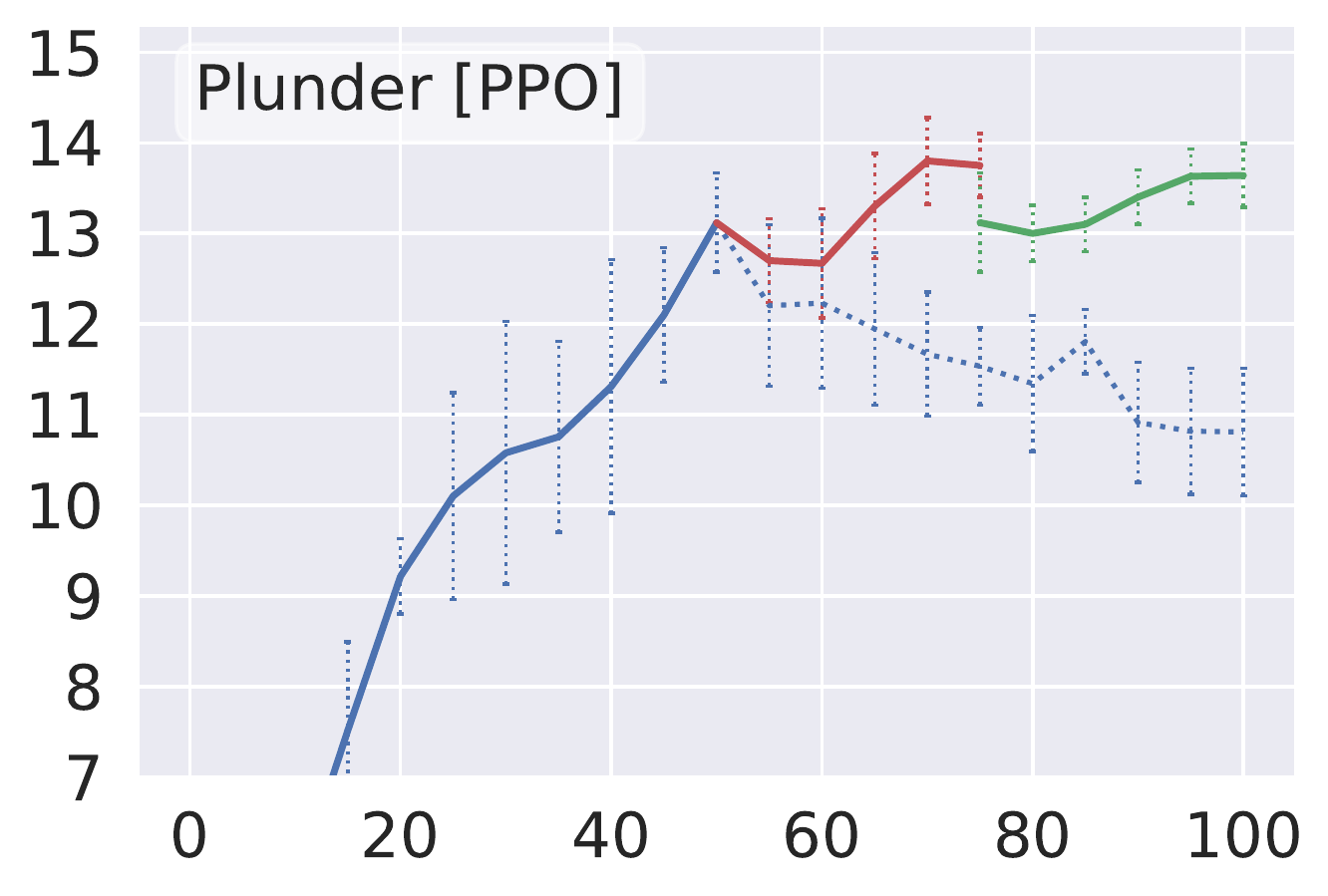}
        \includegraphics[width=0.24\linewidth]{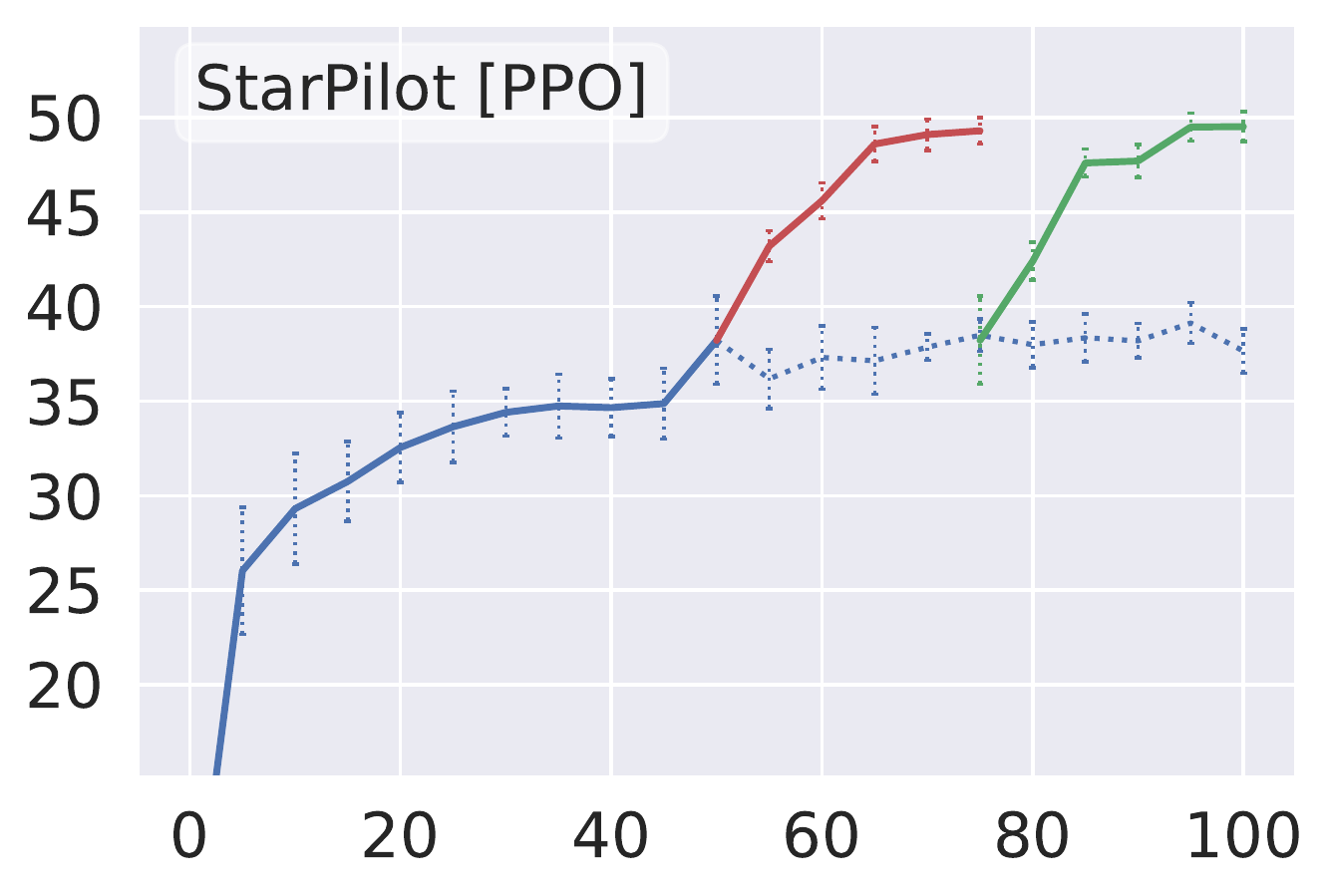}
        \includegraphics[width=0.24\linewidth]{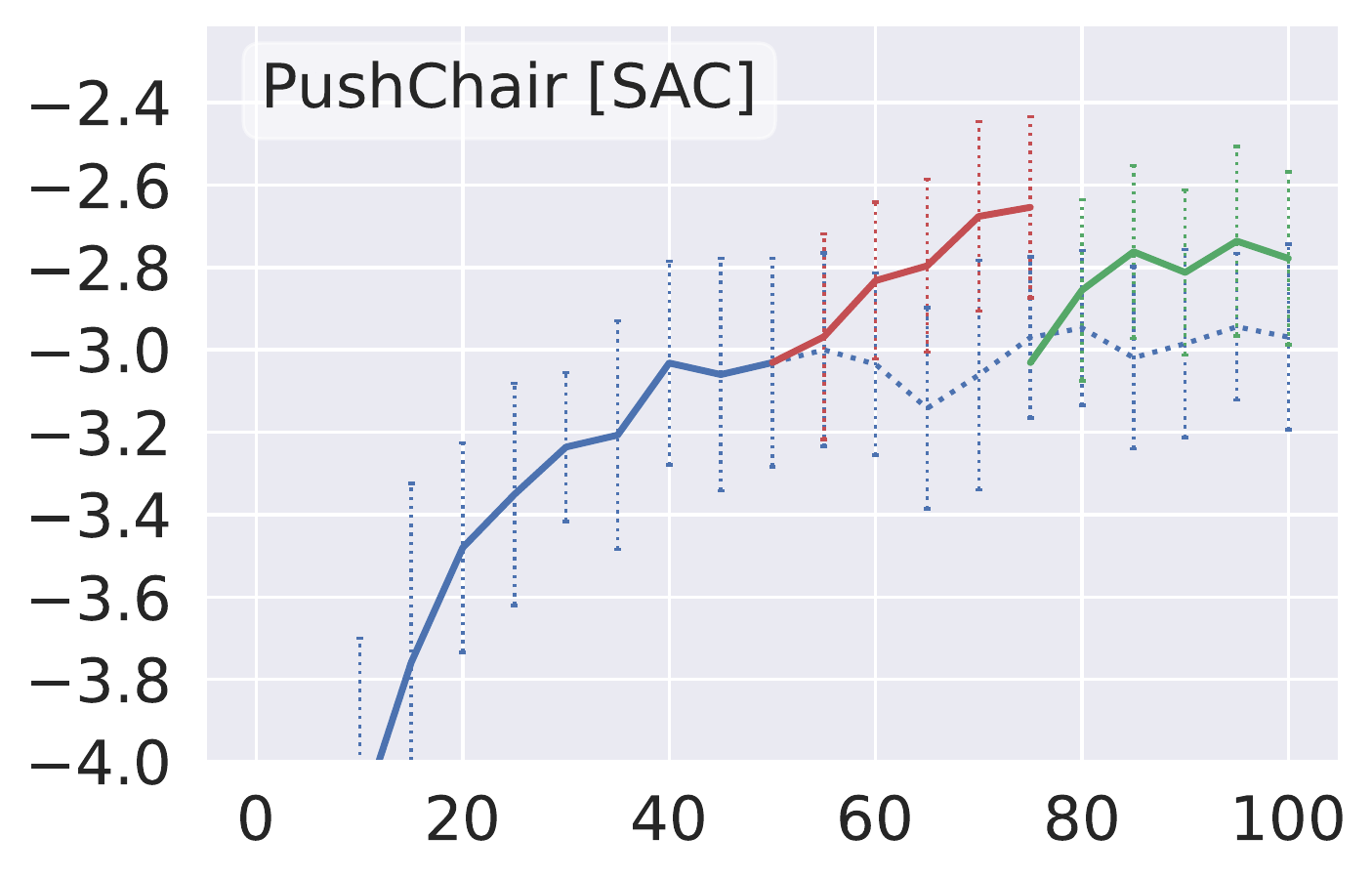}
    \end{subfigure}
    \begin{subfigure}[b]{\linewidth}
        \centering
        \includegraphics[width=0.24\linewidth]{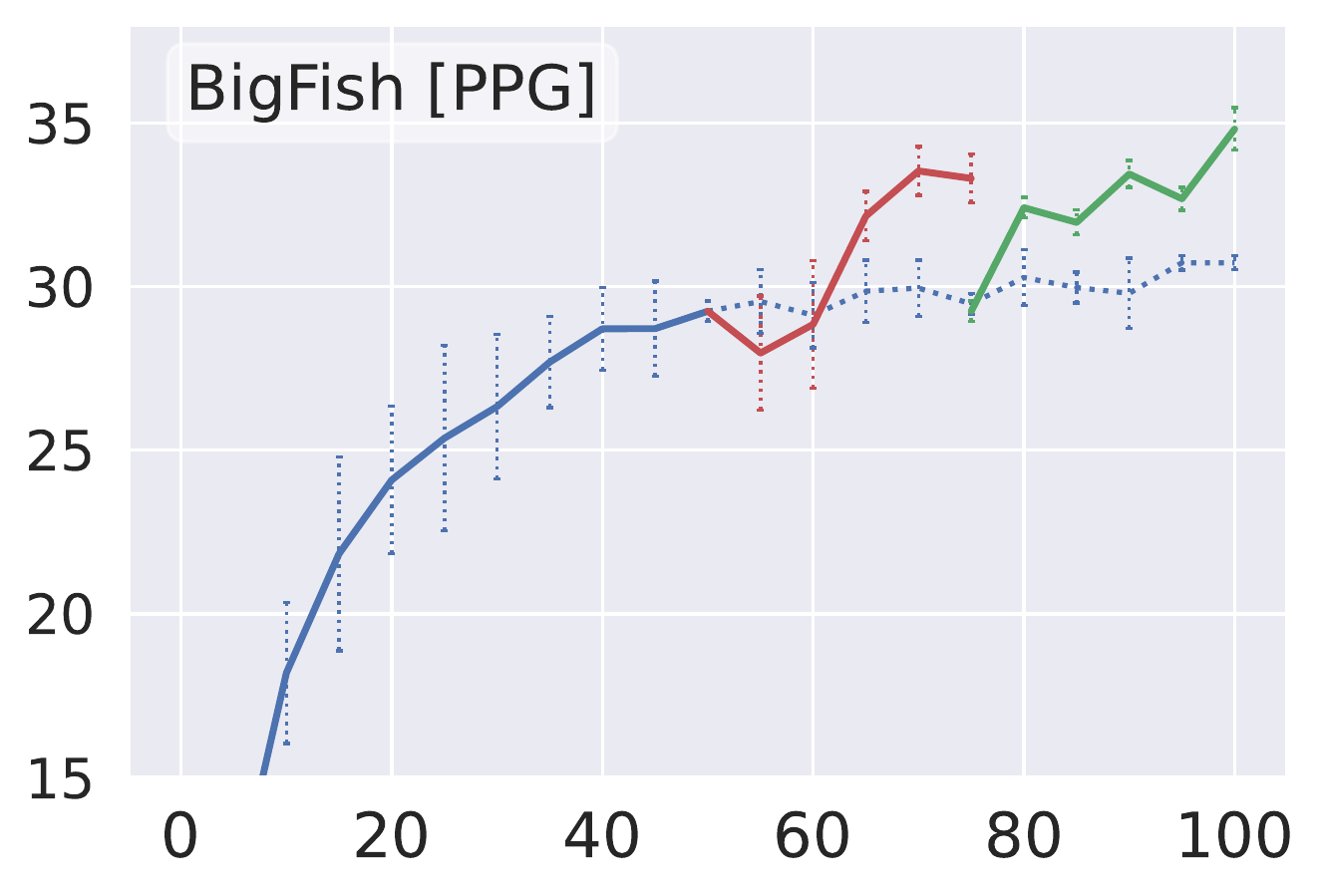}
        \includegraphics[width=0.24\linewidth]{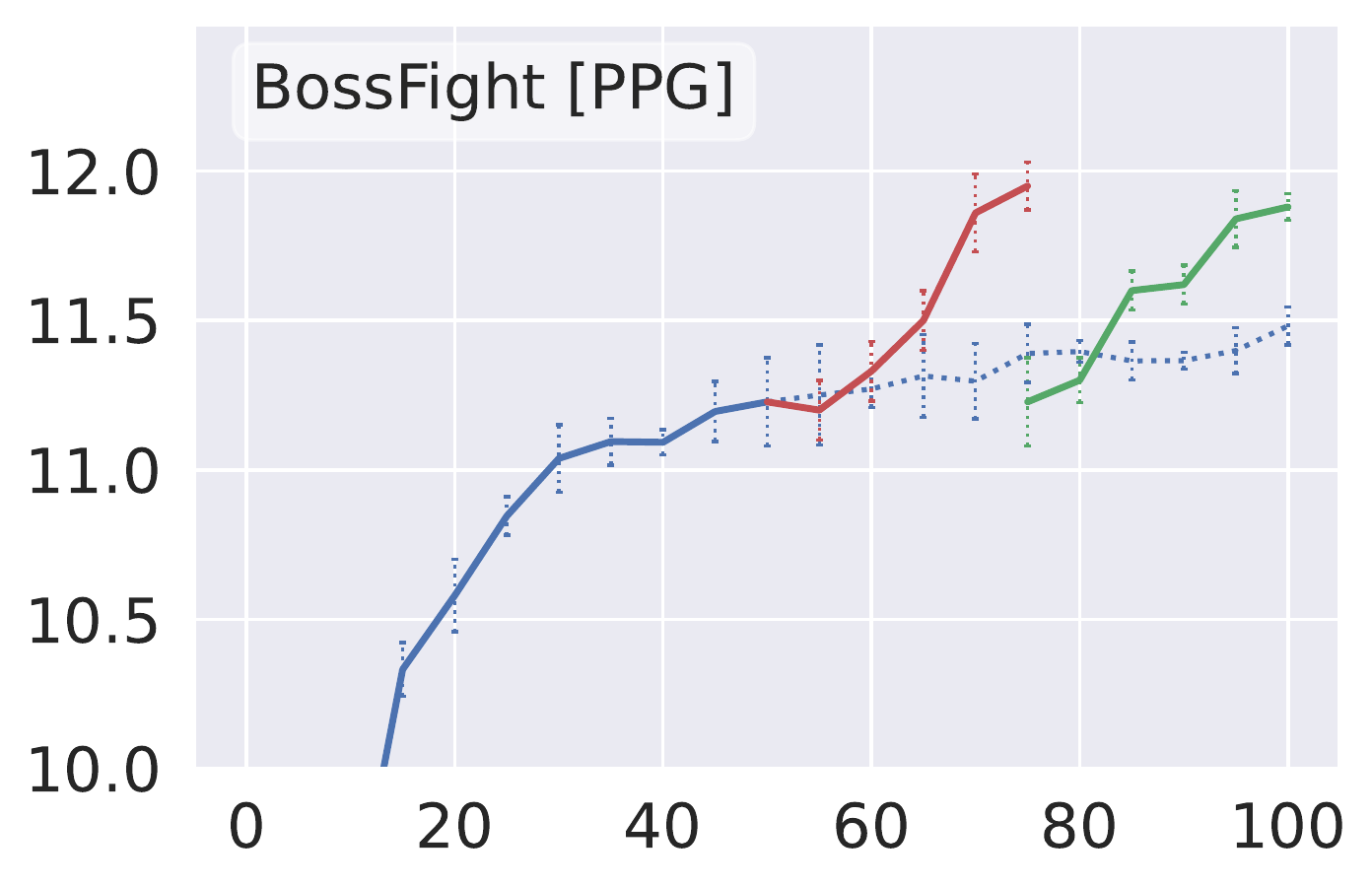}
        \includegraphics[width=0.24\linewidth]{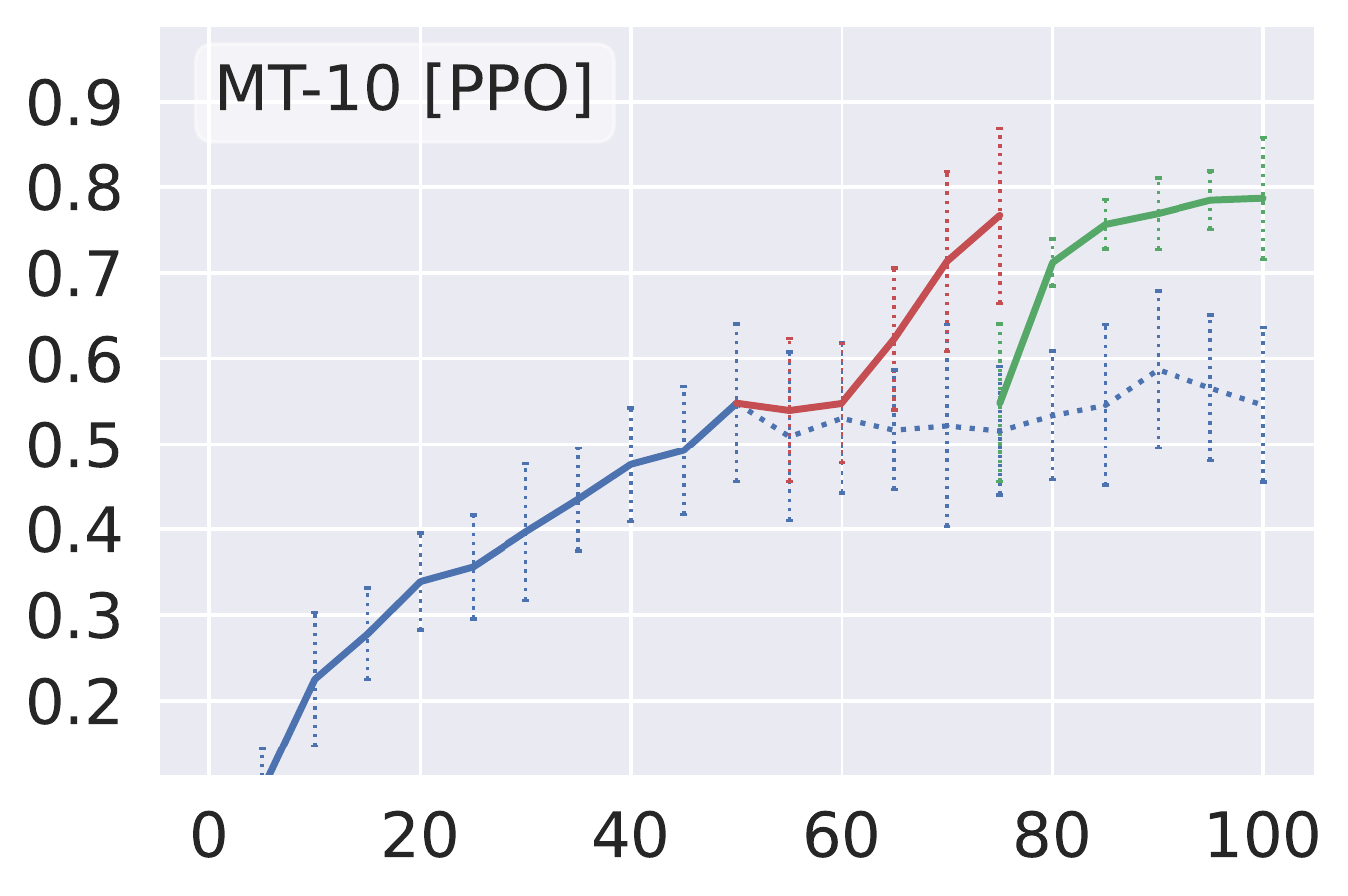}
        \includegraphics[width=0.24\linewidth]{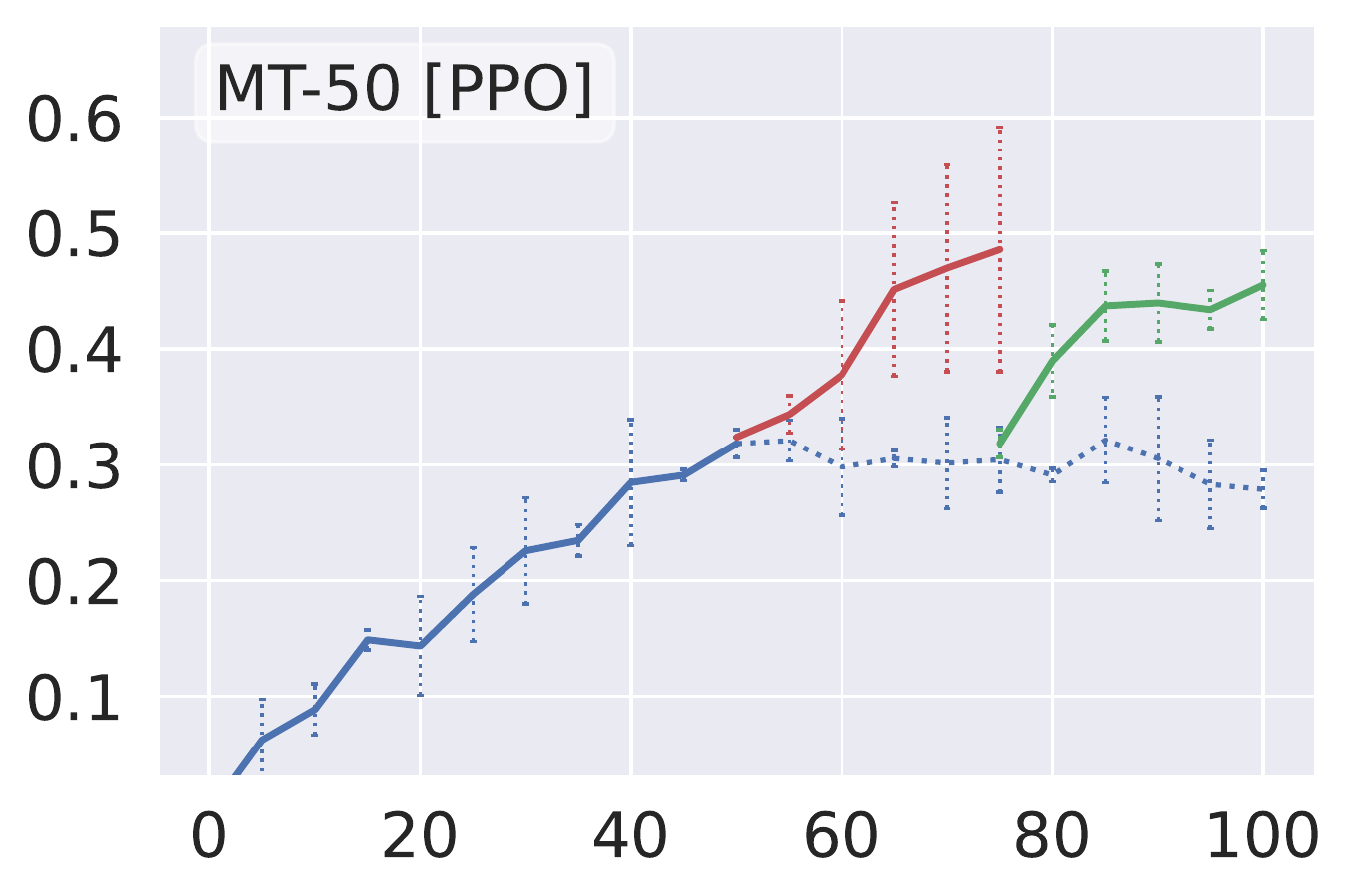}
    \end{subfigure}
\end{subfigure}

\caption{Aggregated training curves for GSL across multiple runs in Procgen, Meta-World and ManiSkill, where GSL consistently improves the baseline (dashed blue curves). The y-axis units are raw episode rewards for Procgen, average success rate for Meta-World and 1000 for ManiSkill. To align the training curves across different runs, we use percentage to represent the x-axis ($0\sim50$\% for initial generalist training, $50\sim75$\% for specialists, and $75\sim100$\% for specialist-guided generalist training).}
\label{fig:training_curves_std}
\end{figure*}